\documentclass[11pt, a4paper, twocolumn, copyright, goog]{google}

\usepackage[authoryear, sort&compress, round]{natbib}
\usepackage{amsmath}
\usepackage{booktabs}
\usepackage{tabularx}
\usepackage{multirow}
\usepackage{graphicx}
\usepackage{amssymb} 
\usepackage{wrapfig}
\usepackage{enumitem}
\usepackage{placeins}
\usepackage[most]{tcolorbox}

\DeclareMathOperator*{\argmax}{arg\,max}
\DeclareMathOperator*{\argmin}{arg\,min}

\uselogo{} 

\title{KeyRec: Bounded Visual Memory for Streaming and Long-Video Understanding}

\reportnumber{ } 

\renewcommand{\today}{}

\author[1,2]{Zihan Chen}
\author[2]{Xuejian Rong}
\author[2]{Xiaojuan Wang}
\author[2]{Boqing Gong}
\author[2]{Adi Zicher}
\author[2]{Yael Pritch}
\author[2]{Nikhil Karnad}

\affil[1]{University of Virginia}
\affil[2]{Google}

\begin{abstract}

Vision-language models are increasingly used to understand long videos and continuous streams. However, dense visual tokens accumulate with video duration, making long-context inference prohibitively expensive. Existing training-free visual-token selection methods reduce this cost by retaining informative tokens, but may lose coherent event evidence and fail to distinguish detailed recent observations from long-range history. We propose \textbf{KeyRec}, a training-free framework for constructing bounded visual memory. During query-agnostic writing, KeyRec preserves fine-grained recent observations in a visual cache and organizes historical evidence into a structured event bank. Candidate events are proposed according to their novelty relative to previously stored events and maintained through an online add--merge--evict update. When a question arrives, a text-only router adaptively allocates a fixed readout budget between recent and event memory, without reprocessing historical frames. KeyRec operates on model-facing visual embeddings and supports both modular encoder--projector VLMs and the encoder- and projector-free NEO-ov architecture. Across four streaming and long-video benchmarks and three VLM backbones, KeyRec achieves the best compressed performance in 13 of 15 settings using only 10\% of the dense decoder-facing visual-token budget. It outperforms the strongest compressed baseline by 2.21--18.37 points on real-time questions, achieves the best compressed result in five of six long-video settings, and performs best in every NEO-ov 2B setting.
\end{abstract}

\begin{document}

\maketitle

\section{Introduction}
\label{sec:introduction}

Recent advances in vision-language models (VLMs) have enabled increasingly capable reasoning over visual content, extending their applications from images and short clips to long-form recordings and continuous video streams~\citep{zhang2024llava,chen2025longvila,zhang2025videollama,cho2026spatialclaw}. However, a VLM typically represents each frame with dozens or hundreds of visual tokens. Consequently, the visual sequence grows linearly with the number of frames, while dense self-attention during prefilling incurs quadratic computation in the sequence length~\citep{ning2025livevlm,wei2025streamvln,zhang2025flash}. This accumulation is particularly problematic for streaming video, whose duration may be unknown and whose queries can arrive at any time~\citep{di2025streaming,yang2025streammem}, as well as for long videos, where dense visual context can exceed the available memory or context budget and may be repeatedly processed for multiple questions~\citep{jin2025efficient,chen2026streamingtom}. Compressing the growing visual context is therefore essential for scalable video understanding~\citep{yang2025visionzip,tao2025dycoke,shao2026holitom}. 

Recent training-free methods exploit redundancy among visual tokens by identifying informative regions through spatial-temporal scoring, retaining tokens that differ from adjacent frames or historical anchors, or maintaining a compact semantic representation of the observed stream~\citep{wang2026accelerating,chen2026streamingtom,song2026towards}. These approaches effectively remove repetitive backgrounds and other low-information content, substantially reducing the visual context. However,  we observe three limitations when the retained representations serve as visual memory. First, novelty is commonly measured against a recent frame, coarse anchors, or a global semantic representation, without explicitly organizing history into temporally localized events. As a result, evolving events may be fragmented across updates, while visually similar but temporally distinct occurrences may be conflated. Second, independently selecting important tokens may discard the supporting context that connects an actor, action, object, and scene into coherent event evidence. Third, existing methods often apply a common retention policy across the video, forcing detailed recent observations and long-range historical evidence to compete under the same policy. Uniformly sparse memory may therefore preserve broad semantics while losing the fine-grained evidence required by questions such as ``What is happening right now?'' These limitations motivate treating compression not only as token selection, but also as bounded visual-memory organization.

To address these challenges, we propose \textbf{KeyRec}, a training-free framework that separates query-agnostic memory construction from query-adaptive, fixed-budget readout. During writing, KeyRec maintains a dense \textit{recent cache} and a structured \textit{key event memory}. Incoming frames propose novel candidate events, which are maintained through an online add--merge--evict update that uses both semantic similarity and temporal proximity to consolidate related observations while keeping the historical memory bounded. When a question arrives, a text-only pass of the same frozen VLM allocates the readout budget between recent and event memory and retrieves historical events by query relevance or temporal coverage, without reprocessing previous frames. Because KeyRec operates on model-facing visual embeddings, the same mechanism applies to both modular encoder--projector VLMs and native one-vision models. We instantiate it on Gemma 4 E2B/E4B~\citep{gemmateam2026gemma4} and NEO-ov 2B~\citep{diao2026pixels}, preserving the temporal and spatial indices of selected NEO-ov tokens. Across four streaming~\citep{niu2025ovo,lin2026streamingbench} and long-video benchmarks~\citep{wu2024longvideobench,fu2026video} and three VLM backbones, KeyRec achieves the best compressed performance in 13 of 15 settings using only 10\% of the dense decoder-facing visual-token budget. Our contributions are summarized as follows:
\begin{itemize}[leftmargin=0.35cm]
    \item We propose KeyRec, a training-free framework for bounded visual memory that decouples query-agnostic memory construction from query-adaptive, fixed-budget readout.  
    \item KeyRec organizes memory into detailed recent evidence and temporally localized historical events, maintained through an online semantic-temporal add--merge--evict update, and supports native one-vision VLMs by preserving token structure.  
    \item Across four benchmarks and three VLM backbones, KeyRec achieves the best compressed result in 13 of 15 settings using only $10\%$ of the dense decoder-facing visual-token budget.
\end{itemize}

\section{Method}
\label{sec:method}

\subsection{Problem Formulation and Overview}
\label{sec:problem_formulation}

\paragraph{Video question answering.}
Consider a video stream $\mathcal{V}=\{x_1,x_2,\ldots\}$.
The model-native visual embedding mechanism $\Phi$ maps each frame $x_t$
to $n$ visual tokens
$H_t=\Phi(x_t)=\{\mathbf{h}_{t,i}\}_{i=1}^{n}$.
A dense VLM answering a query $q$ at time $\tau$ conditions on
$[H_1;\ldots;H_\tau]$. This requires retaining and processing $n\tau$ visual tokens: storage grows linearly with $\tau$, while dense self-attention cost grows quadratically.
Although only streaming video imposes strict causality, both streaming and
long-video understanding benefit from a bounded, query-agnostic memory that
can be reused across queries and read adaptively under a fixed inference
budget.

We formalize this as \emph{query-agnostic memory writing with
query-adaptive, fixed-budget reading}:
\begin{equation}
    \begin{aligned}
        \mathcal{M}_t
        &=
        \operatorname{Update}_{\mathcal{M}}
        (\mathcal{M}_{t-1},H_t),
        &\quad |\mathcal{M}_t| &\leq S, \\
        Z_q
        &=
        \operatorname{Read}(\mathcal{M}_{\tau},q),
        &\quad |Z_q| &\leq B,
    \end{aligned}
    \label{eq:bounded_memory_interface}
\end{equation}
where $S$ is the persistent visual-token storage budget and $B$ is the
per-query readout budget. The answer is generated as
$\hat{y}=\operatorname{VLM}(Z_q,q)$ without revisiting the video.

\begin{figure*}[t]
    \centering
    \includegraphics[width=0.99\textwidth]{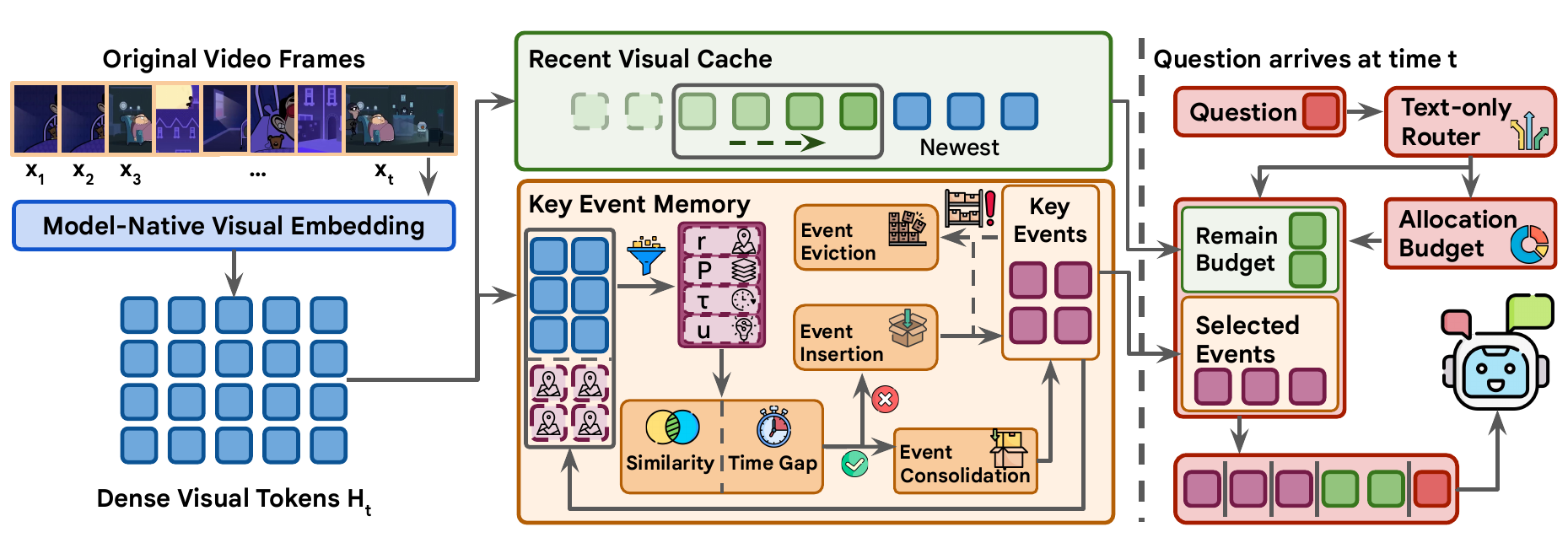} 
    \caption{\textbf{Overview of KeyRec.} During query-agnostic writing, incoming frames update a recent cache and a bounded key-event memory through event insertion, consolidation, and eviction. At query time, a text-only router allocates a fixed readout budget between recent and historical evidence for query-adaptive inference without replaying previous frames.}
    \label{fig:keyrec}\vspace{-0.1in}
\end{figure*}

\paragraph{KeyRec overview.}
KeyRec instantiates this interface as
$\mathcal{M}_t=(\mathcal{C}_t,\mathcal{E}_t)$, where
$\mathcal{C}_t$ is a \emph{recent cache} preserving detailed evidence from the latest
frames and $\mathcal{E}_t=\{e_k\}_{k=1}^{K_t}$ is a bounded \textit{key event memory}
retaining informative historical evidence.
The cache stores at most $N_C$ visual tokens, while the event bank contains at
most $K$ events of at most $m$ tokens each, giving total storage
$|\mathcal{C}_t|+\sum_{e_k\in\mathcal{E}_t}|P_k|
\leq N_C+Km\leq S$.
Both memories are constructed without access to the query.
At query time, a text-only router allocates the fixed budget $B$ between
recent and event memory and selects the evidence presented to the VLM.

\subsection{Query-Agnostic Memory Construction}
\label{sec:memory_construction}

Given a frame $x_t$, KeyRec updates the two memory components using
only its visual tokens $H_t$ and the memory state $\mathcal{M}_{t-1}$ from preceding frames.
\vspace{-0.2in}
\paragraph{Recent visual cache.}

The recent cache $\mathcal{C}_t$ is an ordered sequence containing at most $N_C$ visual tokens from the latest observed frames. It is initialized as $\mathcal{C}_0=[]$ and stores incoming visual tokens. For each incoming frame, the cache appends $H_t$ and retains only the latest $N_C$ tokens:
\begin{equation}
        \mathcal{C}_t =
        \operatorname{Update}_\mathcal{C}
        (\mathcal{C}_{t-1},H_t)=
    \operatorname{Tail}_{N_C}
    \left(\mathcal{C}_{t-1} \mathbin{\Vert} H_t \right),
    \label{eq:recent_cache}
\end{equation}
where $\Vert$ denotes sequence concatenation and $\operatorname{Tail}_{N_C}$ retains the latest $N_C$ tokens. As new frames arrive, older visual tokens are gradually evicted from the cache. Because this fixed-capacity cache only preserves recent evidence, KeyRec complements it with a bounded key-event memory for long-range history.
\vspace{-0.2in}
\paragraph{Key event memory.}

The key event memory is a bounded collection of historical visual evidence,
initialized as $\mathcal{E}_0=\varnothing$.
At time $t$, it contains
\begin{equation}
    \begin{aligned}
        \mathcal{E}_t &= \{e_k\}_{k=1}^{K_t}, \quad K_t\leq K, \\
        e_k &= \left(\mathbf{r}_k, P_k, \boldsymbol{\tau}_k, u_k\right),
    \end{aligned}
    \label{eq:event_memory}
\end{equation}
where $K$ is a fixed event capacity.
For each event, $\mathbf{r}_k\in\mathbb{R}^{d}$ is a compact route embedding,
$P_k\in\mathbb{R}^{m_k\times d}$ is its stored visual tokens with
$m_k\leq m$, $\boldsymbol{\tau}_k$ contains the corresponding token
timestamps, and $u_k$ denotes its event utility.
The key event memory stores at most $Km$ visual tokens.

$\triangleright$\hspace{0.1em}  \textbf{Candidate construction.}
Each incoming frame proposes a candidate event that is not sufficiently represented by the current event memory $\mathcal{E}_{t-1}$.
For each $\mathbf{h}_{t,i}\in H_t$, we define its novelty as
\begin{equation}
    \nu_{t,i} = 1 - \max_{e_k\in\mathcal{E}_{t-1}} \cos(\mathbf{h}_{t,i}, \mathbf{r}_k),
    \label{eq:token_novelty}
\end{equation}
with $\nu_{t,i}=1$ when $\mathcal{E}_{t-1}$ is empty.
$\nu_{t,i}$ measures whether a visual token contributes information not yet
represented by the historical event memory, rather than merely measuring its
within-frame saliency. A larger $\nu_{t,i}$ indicates that the token is less represented by the stored event routes. 

Let $\mathcal{I}_t$ contain the indices of the $m$ most novel tokens in frame $t$: $\mathcal{I}_t = \operatorname{Top-m}(\{\nu_{t,i}\}_{i=1}^{n}),$
we summarize the selected tokens into a candidate route using their normalized novelty scores:
\begin{equation}
    \mathbf{r}_t = 
        \operatorname{Norm}\left(
        \sum_{i\in\mathcal{I}_t} \frac{\exp(\nu_{t,i})}{\sum_{j\in\mathcal{I}_t}\exp(\nu_{t,j})} \mathbf{h}_{t,i}
    \right).
    \label{eq:candidate_route}
\end{equation}
The same selected tokens constitute the candidate event tokens $P_t$, with their timestamps retained in $\boldsymbol{\tau}_t$. We define the candidate utility as their average novelty: $u_t = \sum_{i\in\mathcal{I}_t} \nu_{t,i}/|\mathcal{I}_t|$.
Together, these quantities form the candidate event \(\widetilde{e}_t= (\mathbf{r}_t,P_t,\boldsymbol{\tau}_t,u_t)\), which is subsequently incorporated into the bounded event bank.

$\triangleright$\hspace{0.1em} \textbf{Bounded bank update.}
Each candidate is either inserted as a new event or consolidated with an
existing event; when insertion exceeds the capacity $K$, one stored event is
evicted. To determine whether the candidate continues an existing event, we
first identify the most similar stored route:
\begin{equation}
    s_{t,k}=\cos(\mathbf{r}_t,\mathbf{r}_k),
    \qquad
    k^\star=\argmax_{1\leq k\leq K_{t-1}} s_{t,k}.
    \label{eq:nearest_event}
\end{equation}
Since route similarity alone does not guarantee that events should be merged, as similar visual content can recur across distant timestamps, we additionally constrain the matched event to be temporally adjacent to the candidate. Given that $\mathcal{E}_{t-1}$ contains only events from preceding frames, we define their temporal gap as
\begin{equation}
    g_{t,k^\star}
    =
    \min(\boldsymbol{\tau}_t)
    -
    \max(\boldsymbol{\tau}_{k^\star})
    \geq 0,
    \label{eq:event_time_gap}
\end{equation}
where $\min(\boldsymbol{\tau}_t)$ is the candidate start time and $\max(\boldsymbol{\tau}_{k^\star})$ is the stored-event end time. The candidate is mergeable with $e_{k^\star}$ only if
\begin{equation}
    s_{t,k^\star}\geq\gamma,
    \qquad
    g_{t,k^\star}\leq\delta,
    \label{eq:event_merge_condition}
\end{equation}
where $\gamma$ and $\delta$ control semantic similarity and temporal locality,
respectively. This prevents temporally distant recurring observations from
being collapsed into the same event.

\emph{Candidate insertion.}
If the event bank is empty or Eq.~\eqref{eq:event_merge_condition} is not
satisfied, the candidate is appended as a distinct event,
$\overline{\mathcal{E}}_t=
\mathcal{E}_{t-1}\cup\{\widetilde{e}_t\}$.
If this exceeds capacity $K$, the eviction rule below is applied.

\emph{Event consolidation.}
Otherwise, the candidate is consolidated with $e_{k^\star}$.
Its route is updated by a normalized moving average:
\begin{equation}
    \mathbf{r}_{k^\star}
    \leftarrow
    \operatorname{Norm}
    \left(
        (1-\beta)\mathbf{r}_{k^\star}
        +\beta\mathbf{r}_t
    \right).
    \label{eq:event_route_update}
\end{equation}
The stored visual tokens and timestamps are updated jointly:
\begin{equation}
    \begin{aligned}
        P_{k^\star} &\leftarrow \operatorname{Uniform}_{m}\Big(\operatorname{TimeSort}\big[P_{k^\star}; P_t\big]\Big); \\
        \boldsymbol{\tau}_{k^\star} &\leftarrow \operatorname{Uniform}_{m}\Big(\operatorname{TimeSort}\big[\boldsymbol{\tau}_{k^\star}; \boldsymbol{\tau}_t\big]\Big).        
    \end{aligned}
    \label{eq:payload_merge}
\end{equation}
$\operatorname{TimeSort}$ orders the visual tokens and timestamps by source time. To enforce the per-event storage capacity, $\operatorname{Uniform}_{m}$ leaves sequences with at most $m$ tokens unchanged, while deterministically retaining $m$ uniformly spaced entries. Such an operation preserves the temporal extent of the merged event. Event utility records the strongest candidate novelty and
is updated as
$
u_{k^\star}\leftarrow\max(u_{k^\star},u_t)
$,
which avoids diluting brief distinctive changes with redundant adjacent
observations.

\emph{Capacity-based eviction.}
When the key event bank exceeds $K$ events, KeyRec removes the event with the lowest
retention score, which balances visual utility and temporal coverage.
Let
$
c_k=(\min(\boldsymbol{\tau}_k)+\max(\boldsymbol{\tau}_k))/2
$
be the center time of event $e_k$. We define
\begin{equation}
    d_k^{\mathrm{temp}}
    =
    \min_{j\neq k}|c_k-c_j|,
    \;
    S_{\mathrm{ret}}(e_k)
    =
    u_k+\lambda d_k^{\mathrm{temp}}.
    \label{eq:event_retention}
\end{equation}

The utility term favors events containing distinctive visual changes, while
the temporal term favors events that cover otherwise underrepresented
portions of the video.
KeyRec evicts
$
k_{\mathrm{evict}}=\argmin_k S_{\mathrm{ret}}(e_k)
$
to restore $|\mathcal{E}_t|\leq K$.

\subsection{Query-Adaptive Fixed-Budget Readout}
\label{sec:query_readout}

Different questions require different evidence from the same visual memory: current-state questions favor detailed recent observations, whereas retrospective questions require broader historical coverage. KeyRec therefore uses the same frozen VLM as a text-only memory router.

We define five readout states, ordered from recent- to event-oriented:
\textsc{AllRecent}, \textsc{HeavyRecent}, \textsc{Balanced},
\textsc{HeavyEvent}, and \textsc{AllEvent}, collectively denoted by
$\mathcal{Z}$. Given question $q$, the router predicts
\begin{equation}
    z_q=\operatorname{Router}_{\mathrm{VLM}}(q),
    \qquad z_q\in\mathcal{Z}.
    \label{eq:llm_router}
\end{equation}
The text-only memory router observes only the question and a textual description of the two memory sources; it neither accesses nor reprocesses the video. Each state jointly determines the recent-event allocation and the strategy for selecting historical events.

Let $\pi_q=\pi(z_q)\in[0,1]$ denote the fraction of the readout assigned to event memory, with $\pi_q=0$ and $1$ corresponding to \textsc{AllRecent} and \textsc{AllEvent}, respectively. Given a total visual-token readout budget $B$ and event size $m$, the numbers of selected events and recent tokens are
\begin{equation}
    k_q=
    \min\left\{
        K,
        \operatorname{round}\left(\frac{\pi_q B}{m}\right)
    \right\},
    \;
    r_q=B-k_qm.
    \label{eq:budget_allocation}
\end{equation}

The router state also determines how the $k_q$ historical events are selected.
For \textsc{HeavyRecent} and \textsc{Balanced}, KeyRec uses
relevance-oriented retrieval, selecting the events whose routes have the
highest cosine similarity to the mean-pooled input-token embedding of the question $\mathbf{q}$.
For \textsc{HeavyEvent} and \textsc{AllEvent}, it instead favors temporal
coverage by ordering events by center time and uniformly selecting $k_q$
events across the sequence. \textsc{AllRecent} retrieves no historical events.
Selected events are restored to temporal order before decoding.

The recent contribution is
$C_q=\operatorname{Tail}_{r_q}(\mathcal{C}_{\tau})$.
Together, $C_q$ and the selected event tokens form the fixed-budget readout
$Z_q$, with $|Z_q|\leq B$, which is presented to the vision-language model.

\paragraph{Segmented memory presentation.} We serialize the selected readout $Z_q$ as separate memory segments with neutral numbered labels, preserving their boundaries without assigning semantic event types. The cached VLM-facing embeddings are inserted directly at the corresponding segment positions. Appendix~\ref{sec:memory_serialization_ablation} compares this default serialization with alternative designs.  

\subsection{KeyRec on Native One-Vision Models}
\label{sec:keyrec_neo}

Existing visual-token compression methods have predominantly been developed for modular VLMs with a dedicated vision encoder and visual projector. Native one-vision models such as NEO-ov instead use lightweight patch embeddings and jointly process visual and textual tokens in a unified backbone~\citep{diao2026pixels}, making methods that rely on ViT-specific features or attention signals difficult to transfer directly~\citep{chen2026streamingtom}. We therefore study whether bounded visual memory can be applied to this emerging architecture with minimal adaptation.

Because KeyRec operates directly on model-facing visual embeddings, its memory construction and query-adaptive readout remain unchanged. The only architecture-specific adaptation for NEO-ov is to preserve the positional metadata of each retained token, which we represent as $\xi_i=(\mathbf{h}_i,\tau_i,p_i^h,p_i^w)$, where $\tau_i$ is the source frame and $(p_i^h,p_i^w)$ is its original spatial location. KeyRec operates only on $\mathbf{h}_i$, while the metadata follows the same selection and subsampling indices. At read time, selected tokens are ordered by source time and restored to their native spatial and temporal positions before being inserted into the corresponding \texttt{<IMG\_CONTEXT>} locations. Thus, adapting KeyRec to NEO-ov requires only structure-preserving positional bookkeeping, without modifying the underlying memory algorithm or reconstructing dense visual grids.

\section{Experiments}

\subsection{Experimental Setup}
\label{sec:experimental_setup}

\begin{table*}[t]
\centering
\small
\tabcolsep=5pt
\begin{tabular}{lllc|cccc}
\toprule
Benchmark & Model & Subset & Vanilla &STC-Pruner  &STOM-CTR   &CausalMem   & KeyRec \\
\midrule
OVO-Bench & Gemma 4 E2B & Backward & 43.22 &41.20  & 42.30   &\textbf{45.31} & 43.83 \\
OVO-Bench & Gemma 4 E2B & Realtime & 50.72 &48.02 &  47.23   &55.49 & \textbf{57.70} \\
OVO-Bench & Gemma 4 E4B & Backward & 53.86 &53.01 &50.49     &54.45  & \textbf{54.45}  \\
OVO-Bench & Gemma 4 E4B & Realtime & 59.83 &51.67 &53.26     &62.51  &\textbf{64.81}  \\
OVO-Bench & NEO-ov 2B & Backward & 38.66 & 35.22 &  --  & 39.09   & \textbf{41.90}\\
OVO-Bench & NEO-ov 2B & Realtime & 58.07 & 36.84 &  --  & 44.41   & \textbf{62.78} \\
\midrule
StreamingBench & Gemma 4 E2B & Realtime & 51.28 & 52.91  &51.36   &58.01    & \textbf{61.29} \\
StreamingBench & Gemma 4 E4B & Realtime & 63.73 & 56.44 & 60.25   &63.61    & \textbf{66.49}  \\
StreamingBench & NEO-ov 2B & Realtime & 69.06
& 50.68  & --  &55.56   & \textbf{69.10}  \\
\midrule
LongVideoBench & Gemma 4 E2B & Validation & 35.90 &33.06   & 34.64  & 34.11   & \textbf{35.60} \\
LongVideoBench & Gemma 4 E4B & Validation & 49.14 &43.38   & 43.98  & 43.08    & \textbf{45.77} \\
LongVideoBench & NEO-ov 2B & Validation & 52.51 &44.43   & --  & 42.93   & \textbf{49.29}  \\
\midrule
Video-MME-v2 & Gemma 4 E2B & Overall & 18.47  &14.16   & \textbf{15.69}  & 15.31   & 15.41  \\
Video-MME-v2 & Gemma 4 E4B & Overall & 20.09 &14.34   &15.91  & 14.84   & \textbf{17.63}  \\
Video-MME-v2 & NEO-ov 2B & Overall &22.06 &17.56   & -- & 17.47   & \textbf{20.03} \\
\bottomrule
\end{tabular}
\caption{\textbf{Main results on streaming and long-video understanding benchmarks.} Vanilla retains all visual tokens, whereas compressed methods use a matched decoder-facing budget of approximately $10\%$ of the dense visual input. Bold indicates the best compressed result in each row; ``--'' denotes an unsupported architecture (STOM-CTR: StreamingTOM-CTR).}
\label{tab:main_results}
\end{table*}

\paragraph{Benchmarks and metrics.} 
We evaluate KeyRec on both streaming and long video understanding benchmarks. For streaming evaluation, we use the nine backward-tracing and real-time tasks of OVO-Bench~\citep{niu2025ovo} and the Real-Time Visual Understanding subset of StreamingBench~\citep{lin2026streamingbench}, where each query accesses only its preceding video prefix. For long-video evaluation, we use the visual-only validation split of LongVideoBench~\citep{wu2024longvideobench} and the full Video-MME-v2 benchmark~\citep{fu2026video}. We exclude subtitles and audio to isolate visual-memory compression and report accuracy following each benchmark's official evaluation.

\paragraph{VLM backbones and baselines.}
We evaluate Gemma 4 E2B and E4B as modular encoder--projector VLMs~\citep{gemmateam2026gemma4}, and NEO-ov 2B as a native one-vision model~\citep{diao2026pixels}. We compare against three representative training-free visual-token compression methods: \textbf{STC-Pruner}, which scores tokens against current and historical prototypes~\citep{wang2026accelerating}; \textbf{StreamingTOM-CTR}, which combines adjacent-frame changes with ViT attention~\citep{chen2026streamingtom}; and \textbf{CausalMem}, which maintains a fixed-budget low-rank semantic basis~\citep{song2026towards}. \textbf{Vanilla} retains all sampled visual tokens without compression. StreamingTOM-CTR is evaluated only on Gemma because it requires attention signals from a dedicated vision encoder. All methods use frozen backbones and the same frame-sampling protocol within each setting.

\paragraph{Implementation details.}
For the main experiments, we sample 32 frames from each streaming prefix\footnote{STC-Pruner and StreamingTOM-CTR retain a fixed token fraction per frame, so their retained context grows with stream length. We therefore cap the main comparison at 32 frames to match all methods under the same decoder-facing budget, and separately evaluate fixed-budget full-prefix 1-fps streaming in Section~\ref{sec:fixed_budget_1fps}.} and 64 frames from each complete long video, processing frames chronologically. All compressed methods use approximately $10\%$ of the dense decoder-facing visual-token budget. KeyRec maintains a larger query-agnostic persistent memory of approximately $20\%$ of the dense visual input and selects a $10\%$ readout for each query, whereas the baselines decode their complete retained buffers. Exact memory configurations, router codebooks and prompts, and baseline hyperparameters are provided in Appendix~\ref{sec:additional_implementation}.

\subsection{Main Experimental Results}
\label{sec:main_results}

\begin{figure*}[!t]
    \centering
    \includegraphics[width=0.99\textwidth]{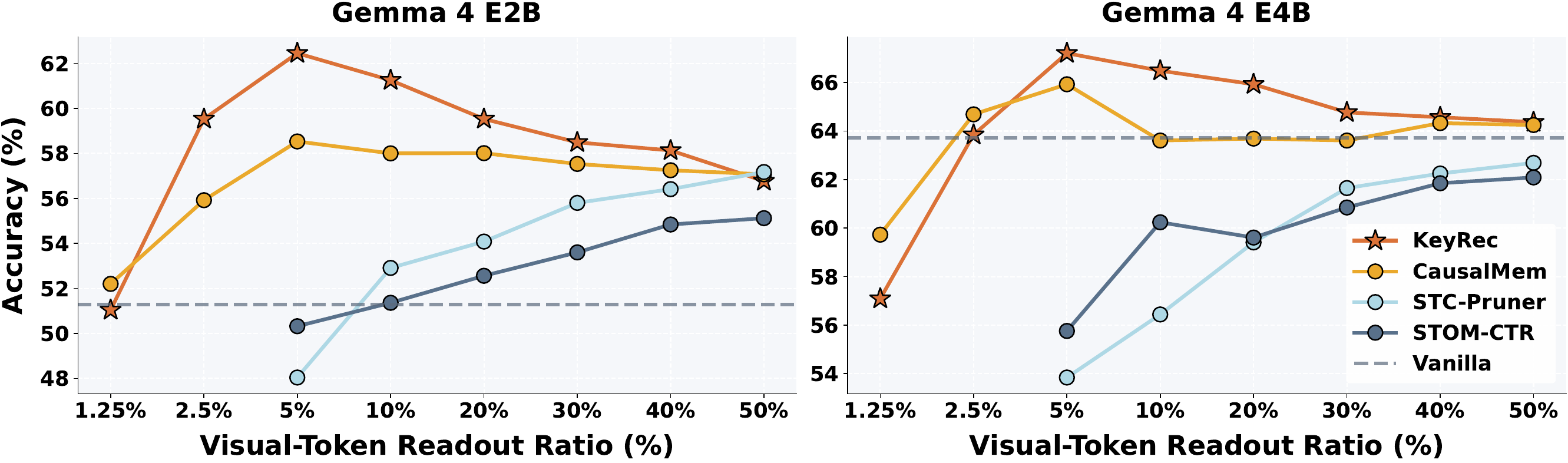} 
    \caption{\textbf{Readout-budget scaling on StreamingBench.} Results are shown for Gemma 4 E2B (left) and Gemma 4 E4B (right). Dashed horizontal lines denote uncompressed Vanilla performance. The horizontal axis reports the decoder-facing visual-token ratio. KeyRec selects each readout from a visual-token memory approximately twice as large, whereas each baseline decodes its complete retained token buffer (STOM-CTR: StreamingTOM-CTR).}
    \label{fig:readout_scaling}
\end{figure*}

Table~\ref{tab:main_results} compares KeyRec with uncompressed inference and representative training-free compression methods under the same frame-sampling protocol.
\textbf{Observation 1: KeyRec is particularly effective for streaming video understanding.} KeyRec achieves the best compressed performance in eight of nine streaming settings and outperforms the strongest compressed baseline by 2.21--18.37 points across OVO-Bench Realtime and StreamingBench. It also matches or exceeds Vanilla in all nine settings, with gains of up to 10.01 points. Although Vanilla retains all visual evidence, it is not an oracle upper bound: redundant streaming prefixes may dilute decisive recent evidence with repeated or irrelevant context. KeyRec instead forms a structured evidence bottleneck in which the event bank compresses redundant history, the recent cache preserves fine-grained latest observations, and the router adapts their allocation to the query.
\textbf{Observation 2: KeyRec provides a strong, but not lossless, long-video compression trade-off.} KeyRec obtains the best compressed result in five of six long-video settings and consistently outperforms the other compressed methods on LongVideoBench. Its 0.30–3.37-point accuracy gap below Vanilla indicates that the fixed 10\% readout preserves much of the model’s long-video capability but does not retain all fine-grained evidence.
\textbf{Observation 3: KeyRec remains effective across substantially different VLM architectures.} KeyRec achieves the best compressed performance in every NEO-ov 2B setting across both streaming and long-video benchmarks. Visual-token compression remains underexplored for encoder- and projector-free one-vision models, whose positional mechanisms and visual interfaces differ from conventional modular VLMs. Our structure-preserving adaptation provides an initial exploration of this setting and suggests the importance of preserving model-native token structure when extending to emerging architectures.

\subsection{Scaling with the Decoder-Facing Readout Ratio}
\label{sec:readout_scaling}

\begin{figure*}[t]
    \centering
    \includegraphics[width=0.99\textwidth]{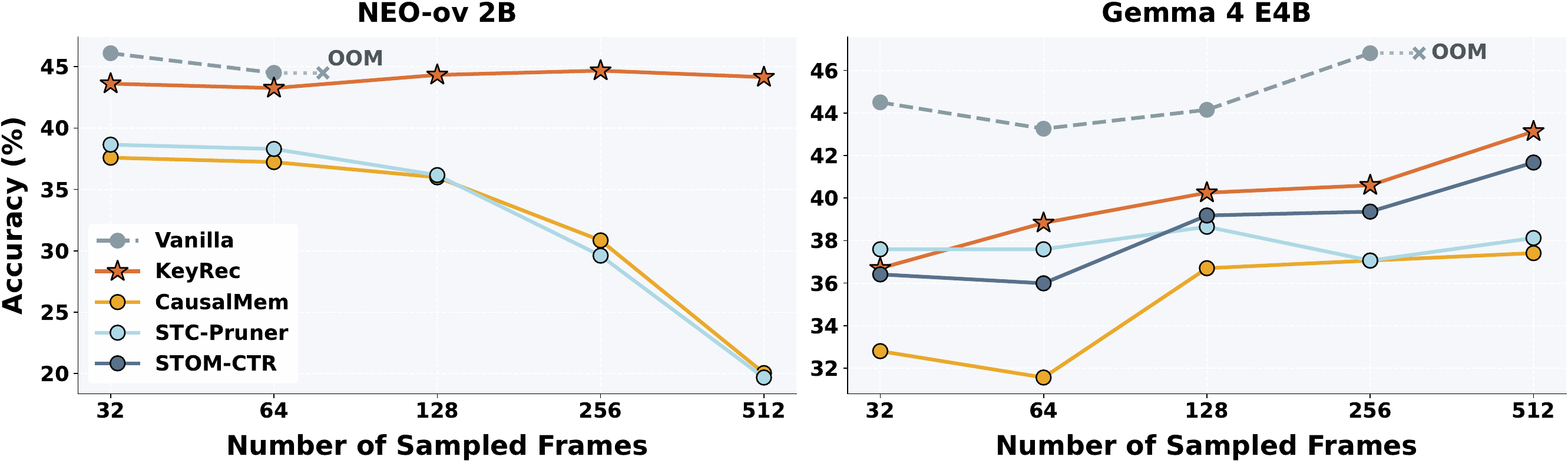} 
    \caption{\textbf{Scaling with the number of sampled frames on LongVideoBench's (900s, 3600s] duration group.} All compressed methods use a $10\%$ decoder-facing ratio. Missing Vanilla results indicate memory limitations; StreamingTOM-CTR (STOM-CTR) does not support NEO-ov.}
    \label{fig:frame_scaling}
\end{figure*}

Figure~\ref{fig:readout_scaling} studies decoder-facing budget scaling on StreamingBench.\footnote{The two smallest ratios stress-test the low-budget regime; at $2.5\%$, the readout contains only 56 tokens, fewer than one dense Gemma frame.} Because KeyRec selects each readout from a persistent memory approximately twice as large, the comparison also controls for its larger candidate pool: At approximately matched persistent storage, KeyRec achieves higher accuracy while exposing only half as many visual tokens to the decoder. Specifically, we compare KeyRec at a 5\% readout with baselines at 10\%, and KeyRec at 10\% with baselines at 20\%. This indicates that its advantage does not arise simply from storing more tokens, but from organizing a structured candidate memory and selecting a query-adaptive readout. 

Across both backbones, KeyRec improves rapidly at small budgets, peaks at a $5\%$ readout, and then gradually declines as the visual context increases. The peak configurations also outperform uncompressed Vanilla, suggesting that additional visual tokens are not uniformly beneficial to a frozen language model: beyond a sufficient evidence budget, redundant or weakly relevant observations may dilute the selected evidence. 

\subsection{Scaling with the Number of Sampled Frames}
\label{sec:frame_scaling}

Figure~\ref{fig:frame_scaling} evaluates scaling on LongVideoBench videos between 900 and 3,600 seconds with a fixed $10\%$ decoder-facing ratio. As more frames are sampled, each method observes denser temporal evidence while receiving a proportionally larger absolute readout budget. Dense Vanilla quickly becomes infeasible, with Gemma running out of memory at 512 frames and NEO-ov at 128 frames or more; where feasible, compressed methods generally remain below Vanilla, indicating that aggressive token reduction does not fully preserve dense long-video performance. The two backbones nevertheless exhibit different scaling behavior: on Gemma E4B, compressed methods generally benefit from additional frames, with KeyRec performing best from 64 frames onward, whereas on NEO-ov, STC-Pruner and CausalMem degrade substantially while KeyRec remains stable across 32--512 frames, staying above $44\%$ accuracy at 512 frames compared with below $25\%$ for both baselines. This divergence suggests that existing token-selection policies may not transfer directly to native one-vision models, where preserving model-native spatial and temporal structure becomes increasingly important over longer videos. KeyRec provides an initial exploration of this setting through lightweight structure-preserving adaptation without modifying the underlying memory algorithm.

\subsection{Effect of Query-Adaptive Memory Allocation}
\label{sec:router_ablation}

\begin{figure}[htbp]
    \centering
    \includegraphics[width=\linewidth]{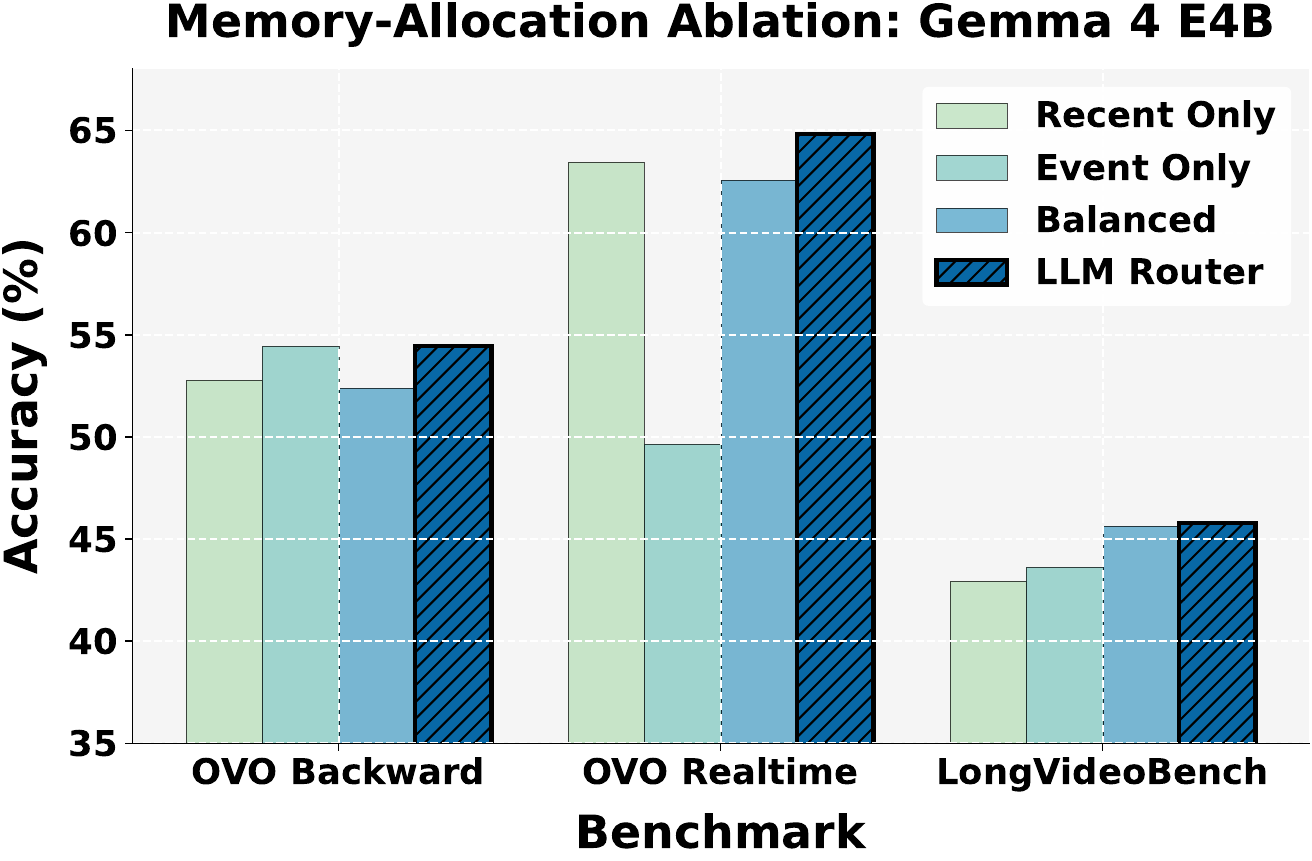} 
    \caption{\textbf{Effect of query-adaptive memory allocation on Gemma 4 E4B.} All policies use the same constructed memory and decoder-facing budget; only the allocation between recent and event memory differs.}
    \label{fig:router_ablation}
\end{figure}


Figure~\ref{fig:router_ablation} compares the query router with three fixed readout policies on Gemma E4B, while keeping the constructed memory and total decoder-facing budget unchanged. The optimal fixed allocation varies substantially across evaluation settings. OVO-Bench Backward benefits most from event memory, whereas OVO-Bench Realtime strongly favors detailed recent evidence and degrades sharply under an event-only readout. LongVideoBench instead performs best with a mixture of recent and historical evidence. Consequently, no single fixed allocation performs consistently well across all three settings. The router adapts the recent--event allocation to each question, matching the event-only policy on backward reasoning while outperforming the fixed policies on real-time and long-video understanding. It therefore achieves the best or tied performance across settings, demonstrating the value of query-adaptive readout over a memory allocation.

\subsection{Latency and Memory Analysis}
\label{sec:efficiency}

\begin{table}[t]
\centering
\small
\tabcolsep=4pt
\renewcommand{\arraystretch}{1.08}
\begin{tabular}{clcc}
\toprule
\textbf{Frames}
& \textbf{Method}
& \textbf{E2E (s)} 
& \textbf{Peak Mem (GiB)} \\
\midrule

\multirow{3}{*}{64}
& Vanilla           & 5.73 & 17.09 \\
& STOM-CTR  & 5.92 & 15.05 \\
& \textbf{KeyRec}   & 7.08 & 15.06 \\
\midrule

\multirow{3}{*}{256}
& Vanilla           & 26.55 & 39.23 \\
& STOM-CTR  & 20.04 & 15.47 \\
& \textbf{KeyRec}   & 22.81 & 16.17 \\
\midrule

\multirow{3}{*}{512}
& Vanilla           & OOM  & OOM\\
& STOM-CTR  & 39.12 & 16.57 \\
& \textbf{KeyRec}   & 43.08 & 18.93 \\
\bottomrule
\end{tabular}
\caption{\textbf{Latency and memory scaling on LongVideoBench with Gemma E4B.} Compressed methods use a $10\%$ decoder-facing ratio.}
\label{tab:efficiency_scaling}
\end{table}

Table~\ref{tab:efficiency_scaling} reports E2E latency and peak GPU memory for Gemma E4B on LongVideoBench, comparing KeyRec with dense Vanilla and StreamingTOM-CTR (STOM-CTR), the strongest compressed competitor at large frame counts. E2E latency is measured from ingestion of the first predecoded frame to the first generated token and includes vision encoding, memory updates, and query processing; peak memory is measured across both ingestion and query processing and includes the model weights.

KeyRec incurs a largely frame-independent overhead from text-only routing, which is progressively amortized as the video grows: relative to STOM-CTR, its E2E overhead decreases from approximately $19\%$ at 64 frames to $10\%$ at 512 frames. KeyRec remains feasible throughout, including at 512 frames, where dense Vanilla runs out of memory; Vanilla already reaches 39.23 GiB at 256 frames. KeyRec uses moderately more memory than STOM-CTR due to its larger candidate memory. Together with the stronger compressed accuracy in Section~\ref{sec:frame_scaling}, these results show that KeyRec trades modest overhead for higher video-understanding performance while scaling substantially better than dense inference.

\subsection{Fixed-Budget 1-FPS Streaming Evaluation}
\label{sec:fixed_budget_1fps}

\begin{table}[t]
\centering
\small
\setlength{\tabcolsep}{8pt}
\renewcommand{\arraystretch}{1.06}
\begin{tabular}{@{}llcc@{}}
\toprule
\textbf{Subset} & \textbf{Method} & \textbf{E2B} & \textbf{E4B} \\
\midrule
\multicolumn{4}{@{}l}{\textit{OVO-Bench}} \\
Backward
& CausalMem & 46.70 & \textbf{56.58} \\
& KeyRec & \textbf{48.61} & 56.15 \\
\addlinespace[2pt]

Realtime
& CausalMem & 57.12 & 62.96 \\
& KeyRec & \textbf{57.49} & \textbf{65.71} \\
\addlinespace[2pt]

\midrule

\multicolumn{4}{@{}l}{\textit{StreamingBench}} \\
Realtime
& CausalMem & 61.20 & 67.77 \\
& KeyRec & \textbf{63.21} & \textbf{68.45} \\
\bottomrule
\end{tabular}
\caption{\textbf{Fixed-budget 1-FPS evaluation on streaming benchmarks with Gemma 4.} Both methods use a fixed decoder-facing budget, independent of the observed prefix length. Bold indicates the better result.}
\label{tab:fixed_budget_1fps}
\end{table}

Our main experiments cap each streaming prefix at 32 sampled frames to enable controlled comparisons with methods whose retained tokens grow with the number of frames. We further evaluate KeyRec under a full-prefix 1-FPS protocol, where all frames available before each query are processed while the visual-memory and decoder-facing budgets remain fixed. We compare KeyRec with CausalMem, which also maintains a globally bounded memory, and omit baselines that retain a fixed token ratio per frame and therefore grow linearly with the observed stream.

The mean video durations of OVO-Bench and StreamingBench are approximately 235 and 264 seconds, respectively. We use 256 frames as a representative reference and set the decoder-facing budget to $B=256\times70\times0.1=1{,}792$ visual tokens for both methods, independent of the actual prefix length. As shown in Table~\ref{tab:fixed_budget_1fps}, KeyRec achieves higher overall performance on both OVO-Bench and StreamingBench across the two backbones, with particularly clear gains on real-time understanding. On OVO-Bench Backward, KeyRec improves E2B and remains close to CausalMem on E4B. Overall, KeyRec maintains its advantage when the 32-frame cap is removed and substantially longer sequences are processed, while keeping persistent storage and per-query decoding cost bounded independently of stream duration.

\paragraph{Efficient Video Understanding}

Efficient video understanding initially focused on compressing a fixed video input by exploiting spatial and temporal redundancy~\citep{shao2026holitom,ren2023testa,bolya2022token,huang2025prunevid,xing2024pyramiddrop,yang2025topv,tao2025dycoke}. Global methods select representative frames, partition videos into coherent segments, or retain a diverse subset of visual tokens~\citep{tang2025adaptive,shen2026fastvid,alvar2025divprune}. More recent approaches make this allocation query-aware, prioritizing frames or regions according to the question~\citep{lin2026videorouter,liu2025keyframe,luo2026quota,zhang2026querystream}, while trained selectors such as AutoGaze remove redundant patches before ViT processing to reduce both vision-encoder and language-model computation~\citep{shi2026attend}. These approaches primarily address which visual evidence should be retained from a fixed or query-conditioned video. KeyRec instead treats the retained evidence as a persistent bounded memory: construction remains
query-agnostic, while query conditioning is deferred to a fixed-budget readout.

\vspace{-0.2in}
\paragraph{Streaming Video Understanding}
Streaming video requires causal processing over an observation horizon that may be unknown~\citep{chen2024videollm,chen2025live,qian2024streaming,wang2026streambridge,qian2025dispider,streamingsurvey}. Training-based systems achieve strong performance through learned response timing, memory organization, and perception--decision coordination, but require model-specific adaptation~\citep{zeng2026streamforest,ding2025streammind,xu2026streamingvlm,zhang2025flash}. Training-free methods instead operate directly on frozen VLMs and have evolved from reducing redundancy in incoming frames to explicitly managing accumulated history~\citep{guan2026video,xiong2025streaming,xie2026fluxmem}. At the visual-token level, STC combines reusable vision features with hierarchical token pruning~\citep{wang2026accelerating}, while StreamingTOM selects tokens using adjacent-frame changes and vision-encoder attention~\citep{chen2026streamingtom}. More memory-oriented approaches explicitly maintain compressed history: FluxMem organizes observations into hierarchical short-, middle-, and long-term memory~\citep{xie2026fluxmem}, while CausalMem maintains a globally bounded semantic basis through online low-rank updates~\citep{song2026towards}. KeyRec follows this causal, training-free trajectory but organizes bounded memory differently: query-agnostic writing separates recent evidence from temporally localized events, which are maintained through semantic-temporal add--merge--evict updates. At query time, a fixed readout budget is adaptively allocated between the two. Complementary work instead manages accumulated language-model KV states through query-agnostic or hierarchical memory~\citep{kim2026infinipot,di2025streaming,yang2025streammem,zhang2026hermes}, whereas KeyRec operates on model-facing visual embeddings before decoding.
\section{Conclusion}
\label{sec:conclusion}

KeyRec constructs bounded visual memory through structured historical events, detailed recent evidence, and query-adaptive readout within a training-free framework. Experiments across streaming and long-video benchmarks show that organizing and presenting visual evidence is as important as selecting individually informative tokens, particularly for questions requiring fine-grained recent context. Our evaluation also reveals an underexplored challenge in adapting visual-token compression to emerging encoder-free VLMs. Although several existing methods can be transferred to these models, their effectiveness may change substantially because architecture-specific signals, positional mechanisms, and visual-token interfaces differ from those of conventional encoder--projector VLMs. As encoder-free architectures such as NEO-ov~\citep{diao2026pixels} and Gemma 4 12B~\citep{gemmateam2026gemma4} continue to emerge, understanding how visual memory should adapt to their native representations constitutes a promising and consequential direction for video understanding.

\bibliography{main}

\begin{thebibliography}{50}
\providecommand{\natexlab}[1]{#1}
\providecommand{\url}[1]{\texttt{#1}}
\expandafter\ifx\csname urlstyle\endcsname\relax
  \providecommand{\doi}[1]{doi: #1}\else
  \providecommand{\doi}{doi: \begingroup \urlstyle{rm}\Url}\fi

\bibitem[Alvar et~al.(2025)Alvar, Singh, Akbari, and Zhang]{alvar2025divprune}
S.~R. Alvar, G.~Singh, M.~Akbari, and Y.~Zhang.
\newblock Divprune: Diversity-based visual token pruning for large multimodal models.
\newblock In \emph{2025 IEEE/CVF Conference on Computer Vision and Pattern Recognition (CVPR)}, pages 9392--9401. IEEE, 2025.

\bibitem[Bolya et~al.(2022)Bolya, Fu, Dai, Zhang, Feichtenhofer, and Hoffman]{bolya2022token}
D.~Bolya, C.-Y. Fu, X.~Dai, P.~Zhang, C.~Feichtenhofer, and J.~Hoffman.
\newblock Token merging: Your vit but faster.
\newblock \emph{arXiv preprint arXiv:2210.09461}, 2022.

\bibitem[Chen et~al.(2024)Chen, Lv, Wu, Lin, Song, Gao, Liu, Gao, Mao, and Shou]{chen2024videollm}
J.~Chen, Z.~Lv, S.~Wu, K.~Q. Lin, C.~Song, D.~Gao, J.-W. Liu, Z.~Gao, D.~Mao, and M.~Z. Shou.
\newblock Videollm-online: Online video large language model for streaming video.
\newblock In \emph{2024 IEEE/CVF Conference on Computer Vision and Pattern Recognition (CVPR)}, pages 18407--18418. IEEE, 2024.

\bibitem[Chen et~al.(2025{\natexlab{a}})Chen, Zeng, Lin, Li, Ma, and Shou]{chen2025live}
J.~Chen, Z.~Zeng, Y.~Lin, W.~Li, Z.~Ma, and M.~Z. Shou.
\newblock Live: Learning video llm with streaming speech transcription at scale.
\newblock In \emph{2025 IEEE/CVF Conference on Computer Vision and Pattern Recognition (CVPR)}, pages 29083--29095. IEEE, 2025{\natexlab{a}}.

\bibitem[Chen et~al.(2026)Chen, Tao, Shao, and Wang]{chen2026streamingtom}
X.~Chen, K.~Tao, K.~Shao, and H.~Wang.
\newblock Streamingtom: Streaming token compression for efficient video understanding.
\newblock In \emph{Proceedings of the IEEE/CVF Conference on Computer Vision and Pattern Recognition}, pages 24675--24685, 2026.

\bibitem[Chen et~al.(2025{\natexlab{b}})Chen, Xue, Li, Hu, Zhu, Li, Fang, Tang, Yang, Liu, et~al.]{chen2025longvila}
Y.~Chen, F.~Xue, D.~Li, Q.~Hu, L.~Zhu, X.~Li, Y.~Fang, H.~Tang, S.~Yang, Z.~Liu, et~al.
\newblock Longvila: Scaling long-context visual language models for long videos.
\newblock In \emph{International Conference on Learning Representations}, volume 2025, pages 18227--18246, 2025{\natexlab{b}}.

\bibitem[Cho et~al.(2026)Cho, Hachiuma, Badki, Su, Lee, Song, Liu, Radhakrishnan, Kim, Wang, et~al.]{cho2026spatialclaw}
S.~Cho, R.~Hachiuma, A.~Badki, H.~Su, B.-K. Lee, C.~H. Song, S.~Liu, S.~Radhakrishnan, S.~Kim, Y.-C.~F. Wang, et~al.
\newblock Spatialclaw: Rethinking action interface for agentic spatial reasoning.
\newblock \emph{arXiv preprint arXiv:2606.13673}, 2026.

\bibitem[Di et~al.(2025)Di, Yu, Zhang, Li, Cheng, Li, He, Shu, and Jiang]{di2025streaming}
S.~Di, Z.~Yu, G.~Zhang, H.~Li, H.~Cheng, B.~Li, W.~He, F.~Shu, and H.~Jiang.
\newblock Streaming video question-answering with in-context video kv-cache retrieval.
\newblock In \emph{International Conference on Learning Representations}, volume 2025, pages 42115--42127, 2025.

\bibitem[Diao et~al.(2026)Diao, Wang, Wu, Dong, Niu, Zhu, Cai, Fan, Dai, Wu, et~al.]{diao2026pixels}
H.~Diao, J.~Wang, P.~Wu, Y.~Dong, Y.~Niu, Y.~Zhu, Z.~Cai, W.~Fan, L.~Dai, S.~Wu, et~al.
\newblock From pixels to words--towards native one-vision models at scale.
\newblock \emph{arXiv preprint arXiv:2605.28820}, 2026.

\bibitem[Ding et~al.(2025)Ding, Wu, Yang, Jiang, Zhang, Bai, Chen, and Cao]{ding2025streammind}
X.~Ding, H.~Wu, Y.~Yang, S.~Jiang, Q.~Zhang, D.~Bai, Z.~Chen, and T.~Cao.
\newblock Streammind: Unlocking full frame rate streaming video dialogue through event-gated cognition.
\newblock In \emph{2025 IEEE/CVF International Conference on Computer Vision (ICCV)}, pages 13448--13459. IEEE, 2025.

\bibitem[Fu et~al.(2026)Fu, Yuan, Dong, Zhang, Shen, Hu, Li, Su, Long, Xie, et~al.]{fu2026video}
C.~Fu, H.~Yuan, Y.~Dong, Y.-F. Zhang, Y.~Shen, X.~Hu, X.~Li, J.~Su, C.~Long, X.~Xie, et~al.
\newblock Video-mme-v2: Towards the next stage in benchmarks for comprehensive video understanding.
\newblock \emph{arXiv preprint arXiv:2604.05015}, 2026.

\bibitem[{Gemma Team}(2026)]{gemmateam2026gemma4}
{Gemma Team}.
\newblock Gemma 4 technical report, 2026.
\newblock URL \url{https://arxiv.org/abs/2607.02770}.

\bibitem[Guan et~al.(2026)Guan, Yin, Liang, Ju, Luo, Luan, Liu, and Bai]{guan2026video}
Y.~Guan, L.~Yin, D.~Liang, J.~Ju, Z.~Luo, J.~Luan, Y.~Liu, and X.~Bai.
\newblock Video streaming thinking: Videollms can watch and think simultaneously.
\newblock \emph{arXiv preprint arXiv:2603.12262}, 2026.

\bibitem[Huang et~al.(2025)Huang, Zhou, and Han]{huang2025prunevid}
X.~Huang, H.~Zhou, and K.~Han.
\newblock Prunevid: Visual token pruning for efficient video large language models.
\newblock In \emph{Findings of the Association for Computational Linguistics: ACL 2025}, pages 19959--19973, 2025.

\bibitem[Jin et~al.(2025)Jin, Li, Gu, Liu, Zhao, Lai, Gan, Wang, Wang, Tan, et~al.]{jin2025efficient}
Y.~Jin, J.~Li, T.~Gu, Y.~Liu, B.~Zhao, J.~Lai, Z.~Gan, Y.~Wang, C.~Wang, X.~Tan, et~al.
\newblock Efficient multimodal large language models: A survey.
\newblock \emph{Visual Intelligence}, 3\penalty0 (1):\penalty0 27, 2025.

\bibitem[Kim et~al.(2026{\natexlab{a}})Kim, Parthasarathy, Qin, Hur, Sun, Han, Yang, and Gong]{kim2026liteframe}
J.~Kim, N.~Parthasarathy, D.~Qin, J.~Hur, D.~Sun, B.~Han, M.-H. Yang, and B.~Gong.
\newblock Liteframe: Efficient vision encoders unlock frame scaling in video llms.
\newblock \emph{arXiv preprint arXiv:2605.17260}, 2026{\natexlab{a}}.

\bibitem[Kim et~al.(2026{\natexlab{b}})Kim, Shim, Choi, and Chang]{kim2026infinipot}
M.~Kim, K.~Shim, J.~Choi, and S.~Chang.
\newblock Infinipot-v: Memory-constrained kv cache compression for streaming video understanding.
\newblock \emph{Advances in Neural Information Processing Systems}, 38:\penalty0 138983--139013, 2026{\natexlab{b}}.

\bibitem[Lin et~al.(2026{\natexlab{a}})Lin, Fang, Chen, Cheng, Wan, Luo, Wang, Li, Liu, and Sun]{lin2026streamingbench}
J.~Lin, Z.~Fang, C.~Chen, H.~Cheng, Z.~Wan, F.~Luo, Z.~Wang, P.~Li, Y.~Liu, and M.~Sun.
\newblock Streamingbench: Assessing the gap for mllms to achieve streaming video understanding.
\newblock In \emph{ICASSP 2026-2026 IEEE International Conference on Acoustics, Speech and Signal Processing (ICASSP)}, pages 12147--12151. IEEE, 2026{\natexlab{a}}.

\bibitem[Lin et~al.(2026{\natexlab{b}})Lin, Zhang, and Li]{lin2026videorouter}
K.~Lin, W.~Zhang, and G.~Li.
\newblock Videorouter: Query-adaptive dual routing for efficient long-video understanding.
\newblock \emph{arXiv preprint arXiv:2605.05848}, 2026{\natexlab{b}}.

\bibitem[Liu et~al.(2025)Liu, Sun, Lin, Zhang, Zhang, Yin, Wang, Li, and Chen]{liu2025keyframe}
Y.~Liu, J.~Sun, Y.~Lin, J.~Zhang, J.~Zhang, M.~Yin, Q.~Wang, H.~Li, and Y.~Chen.
\newblock Keyframe-oriented vision token pruning: Enhancing efficiency of large vision language models on long-form video processing.
\newblock In \emph{2025 IEEE/CVF International Conference on Computer Vision (ICCV)}, pages 20802--20811. IEEE, 2025.

\bibitem[Luo et~al.(2026)Luo, Chen, Huang, Yin, Lin, Huang, Fu, Ji, Zheng, and Luo]{luo2026quota}
Y.~Luo, W.~Chen, W.~Huang, S.~Yin, H.~Lin, J.~Huang, C.~Fu, J.~Ji, X.~Zheng, and J.~Luo.
\newblock Quota: Query-oriented token assignment via cot query decouple for long video comprehension.
\newblock In \emph{Proceedings of the AAAI Conference on Artificial Intelligence}, volume~40, pages 24160--24168, 2026.

\bibitem[Ning et~al.(2025)Ning, Liu, Jin, Li, Ding, Guo, and Zhao]{ning2025livevlm}
Z.~Ning, G.~Liu, Q.~Jin, C.~Li, W.~Ding, M.~Guo, and J.~Zhao.
\newblock Livevlm: Efficient online video understanding via streaming-oriented kv cache and retrieval.
\newblock \emph{arXiv preprint arXiv:2505.15269}, 2025.

\bibitem[Niu et~al.(2025)Niu, Li, Miao, Ge, Zhou, He, Dong, Duan, Ding, Qian, et~al.]{niu2025ovo}
J.~Niu, Y.~Li, Z.~Miao, C.~Ge, Y.~Zhou, Q.~He, X.~Dong, H.~Duan, S.~Ding, R.~Qian, et~al.
\newblock Ovo-bench: How far is your video-llms from real-world online video understanding?
\newblock In \emph{2025 IEEE/CVF Conference on Computer Vision and Pattern Recognition (CVPR)}, pages 18902--18913. IEEE, 2025.

\bibitem[Qian et~al.(2024)Qian, Dong, Zhang, Zang, Ding, Lin, and Wang]{qian2024streaming}
R.~Qian, X.~Dong, P.~Zhang, Y.~Zang, S.~Ding, D.~Lin, and J.~Wang.
\newblock Streaming long video understanding with large language models.
\newblock \emph{Advances in Neural Information Processing Systems}, 37:\penalty0 119336--119360, 2024.

\bibitem[Qian et~al.(2025)Qian, Ding, Dong, Zhang, Zang, Cao, Lin, and Wang]{qian2025dispider}
R.~Qian, S.~Ding, X.~Dong, P.~Zhang, Y.~Zang, Y.~Cao, D.~Lin, and J.~Wang.
\newblock Dispider: Enabling video llms with active real-time interaction via disentangled perception, decision, and reaction.
\newblock In \emph{2025 IEEE/CVF Conference on Computer Vision and Pattern Recognition (CVPR)}, pages 24045--24055. IEEE, 2025.

\bibitem[Ren et~al.(2023)Ren, Chen, Li, Sun, and Hou]{ren2023testa}
S.~Ren, S.~Chen, S.~Li, X.~Sun, and L.~Hou.
\newblock Testa: Temporal-spatial token aggregation for long-form video-language understanding.
\newblock In \emph{Findings of the Association for Computational Linguistics: EMNLP 2023}, pages 932--947, 2023.

\bibitem[Shao et~al.(2026)Shao, Tao, Qin, You, Sui, and Wang]{shao2026holitom}
K.~Shao, K.~Tao, C.~Qin, H.~You, Y.~Sui, and H.~Wang.
\newblock Holitom: Holistic token merging for fast video large language models.
\newblock \emph{Advances in Neural Information Processing Systems}, 38:\penalty0 135547--135570, 2026.

\bibitem[Shen et~al.(2026)Shen, Gong, He, Zhang, Liu, Zhao, et~al.]{shen2026fastvid}
L.~Shen, G.~Gong, T.~He, Y.~Zhang, P.~Liu, S.~Zhao, et~al.
\newblock Fastvid: Dynamic density pruning for fast video large language models.
\newblock \emph{Advances in Neural Information Processing Systems}, 38:\penalty0 123553--123581, 2026.

\bibitem[Shi et~al.(2026)Shi, Fu, Lian, Ye, Eigen, Reite, Kautz, Li, Chan, Darrell, et~al.]{shi2026attend}
B.~Shi, S.~Fu, L.~Lian, H.~Ye, D.~Eigen, A.~Reite, J.~Kautz, B.~Li, D.~M. Chan, T.~Darrell, et~al.
\newblock Attend before attention: Efficient and scalable video understanding via autoregressive gazing.
\newblock In \emph{Proceedings of the IEEE/CVF Conference on Computer Vision and Pattern Recognition}, pages 17022--17034, 2026.

\bibitem[Song et~al.(2026)Song, Lin, Wu, Chen, Peng, Chen, Zhou, and Ji]{song2026towards}
B.~Song, Y.~Lin, Q.~Wu, T.~Chen, J.~Peng, X.~Chen, Y.~Zhou, and R.~Ji.
\newblock Towards a dynamic and fixed-budget memory bank for efficient streaming video understanding.
\newblock \emph{arXiv preprint arXiv:2606.25658}, 2026.

\bibitem[Tang et~al.(2025)Tang, Qiu, Xie, Tian, Jiao, and Ye]{tang2025adaptive}
X.~Tang, J.~Qiu, L.~Xie, Y.~Tian, J.~Jiao, and Q.~Ye.
\newblock Adaptive keyframe sampling for long video understanding.
\newblock In \emph{2025 IEEE/CVF Conference on Computer Vision and Pattern Recognition (CVPR)}, pages 29118--29128. IEEE, 2025.

\bibitem[Tao et~al.(2025)Tao, Qin, You, Sui, and Wang]{tao2025dycoke}
K.~Tao, C.~Qin, H.~You, Y.~Sui, and H.~Wang.
\newblock Dycoke: Dynamic compression of tokens for fast video large language models.
\newblock In \emph{2025 IEEE/CVF Conference on Computer Vision and Pattern Recognition (CVPR)}, pages 18992--19001. IEEE, 2025.

\bibitem[Wang et~al.(2026{\natexlab{a}})Wang, Feng, Lai, Xu, Li, Ge, Dehghan, Cao, and Huang]{wang2026streambridge}
H.~Wang, B.~Feng, Z.~Lai, M.~Xu, S.~Li, W.~Ge, A.~Dehghan, M.~Cao, and P.~Huang.
\newblock Streambridge: Turning your offline video large language model into a proactive streaming assistant.
\newblock \emph{Advances in Neural Information Processing Systems}, 38:\penalty0 132332--132359, 2026{\natexlab{a}}.

\bibitem[Wang et~al.(2026{\natexlab{b}})Wang, Liu, Gui, Lin, Yang, Liao, Chen, and Zhang]{wang2026accelerating}
Y.~Wang, X.~Liu, X.~Gui, X.~Lin, B.~Yang, C.~Liao, T.~Chen, and L.~Zhang.
\newblock Accelerating streaming video large language models via hierarchical token compression.
\newblock In \emph{Proceedings of the IEEE/CVF Conference on Computer Vision and Pattern Recognition}, pages 18523--18533, 2026{\natexlab{b}}.

\bibitem[Wei et~al.(2025)Wei, Wan, Yu, Wang, Yang, Mao, Zhu, Cai, Wang, Chen, et~al.]{wei2025streamvln}
M.~Wei, C.~Wan, X.~Yu, T.~Wang, Y.~Yang, X.~Mao, C.~Zhu, W.~Cai, H.~Wang, Y.~Chen, et~al.
\newblock Streamvln: Streaming vision-and-language navigation via slowfast context modeling.
\newblock \emph{arXiv preprint arXiv:2507.05240}, 2025.

\bibitem[Wu et~al.(2024)Wu, Li, Chen, and Li]{wu2024longvideobench}
H.~Wu, D.~Li, B.~Chen, and J.~Li.
\newblock Longvideobench: A benchmark for long-context interleaved video-language understanding.
\newblock \emph{Advances in Neural Information Processing Systems}, 37:\penalty0 28828--28857, 2024.

\bibitem[Xie et~al.(2026)Xie, He, Wang, Zheng, Ye, and Wu]{xie2026fluxmem}
Y.~Xie, B.~He, J.~Wang, X.~Zheng, Z.~Ye, and Z.~Wu.
\newblock Fluxmem: Adaptive hierarchical memory for streaming video understanding.
\newblock In \emph{CVPR}, 2026.

\bibitem[Xing et~al.(2024)Xing, Huang, Dong, Lu, Zhang, Zang, Cao, He, Wang, Wu, et~al.]{xing2024pyramiddrop}
L.~Xing, Q.~Huang, X.~Dong, J.~Lu, P.~Zhang, Y.~Zang, Y.~Cao, C.~He, J.~Wang, F.~Wu, et~al.
\newblock Pyramiddrop: Accelerating your large vision-language models via pyramid visual redundancy reduction.
\newblock \emph{arXiv preprint arXiv:2410.17247}, 2024.

\bibitem[Xiong et~al.(2025)Xiong, Yang, Yu, Zhuge, Zhang, Zhu, and Lu]{xiong2025streaming}
H.~Xiong, Z.~Yang, J.~Yu, Y.~Zhuge, L.~Zhang, J.~Zhu, and H.~Lu.
\newblock Streaming video understanding and multi-round interaction with memory-enhanced knowledge.
\newblock In \emph{International Conference on Learning Representations}, volume 2025, pages 69332--69351, 2025.

\bibitem[Xu et~al.(2026)Xu, Xiao, Chen, He, Peng, Lu, and Han]{xu2026streamingvlm}
R.~Xu, G.~Xiao, Y.~Chen, L.~He, K.~Peng, Y.~Lu, and S.~Han.
\newblock Streamingvlm: Real-time understanding for infinite video streams.
\newblock In \emph{International Conference on Learning Representations}, volume 2026, pages 61463--61475, 2026.

\bibitem[Yang et~al.(2025{\natexlab{a}})Yang, Sui, Xiao, Huang, Gong, Li, Yan, Bai, Sadayappan, Hu, et~al.]{yang2025topv}
C.~Yang, Y.~Sui, J.~Xiao, L.~Huang, Y.~Gong, C.~Li, J.~Yan, Y.~Bai, P.~Sadayappan, X.~Hu, et~al.
\newblock Topv: Compatible token pruning with inference time optimization for fast and low-memory multimodal vision language model.
\newblock In \emph{2025 IEEE/CVF Conference on Computer Vision and Pattern Recognition (CVPR)}, pages 19803--19813. IEEE, 2025{\natexlab{a}}.

\bibitem[Yang et~al.(2025{\natexlab{b}})Yang, Chen, Tian, Wang, Li, Yu, and Jia]{yang2025visionzip}
S.~Yang, Y.~Chen, Z.~Tian, C.~Wang, J.~Li, B.~Yu, and J.~Jia.
\newblock Visionzip: Longer is better but not necessary in vision language models.
\newblock In \emph{2025 IEEE/CVF Conference on Computer Vision and Pattern Recognition (CVPR)}, pages 19792--19802. IEEE, 2025{\natexlab{b}}.

\bibitem[Yang et~al.(2025{\natexlab{c}})Yang, Zhao, Shukla, Singh, Mishra, Zhang, and Ren]{yang2025streammem}
Y.~Yang, Z.~Zhao, S.~N. Shukla, A.~Singh, S.~K. Mishra, L.~Zhang, and M.~Ren.
\newblock Streammem: Query-agnostic kv cache memory for streaming video understanding.
\newblock \emph{arXiv preprint arXiv:2508.15717}, 2025{\natexlab{c}}.

\bibitem[Yang et~al.(2026)Yang, Zhang, Liu, Liu, Ying, Xue, Hou, Qian, and Xu]{streamingsurvey}
Z.~Yang, K.~Zhang, Q.~Liu, T.~Liu, L.~Ying, D.~Xue, Q.~Hou, S.~Qian, and C.~Xu.
\newblock Towards online interactors: A comprehensive survey on streaming video understanding.
\newblock \emph{Preprints}, June 2026.

\bibitem[Zeng et~al.(2026)Zeng, Qiu, Zhang, Li, Wang, Li, Yan, Tian, Tian, Zhao, et~al.]{zeng2026streamforest}
X.~Zeng, K.~Qiu, Q.~Zhang, X.~Li, J.~Wang, J.~Li, Z.~Yan, K.~Tian, M.~Tian, X.~Zhao, et~al.
\newblock Streamforest: Efficient online video understanding with persistent event memory.
\newblock \emph{Advances in Neural Information Processing Systems}, 38:\penalty0 75804--75835, 2026.

\bibitem[Zhang et~al.(2025{\natexlab{a}})Zhang, Li, Cheng, Hu, Yuan, Chen, Leng, Jiang, Zhang, Li, et~al.]{zhang2025videollama}
B.~Zhang, K.~Li, Z.~Cheng, Z.~Hu, Y.~Yuan, G.~Chen, S.~Leng, Y.~Jiang, H.~Zhang, X.~Li, et~al.
\newblock Videollama 3: Frontier multimodal foundation models for image and video understanding.
\newblock \emph{arXiv preprint arXiv:2501.13106}, 2025{\natexlab{a}}.

\bibitem[Zhang et~al.(2025{\natexlab{b}})Zhang, Wang, Tang, Liu, Feng, and Jin]{zhang2025flash}
H.~Zhang, Y.~Wang, Y.~Tang, Y.~Liu, J.~Feng, and X.~Jin.
\newblock Flash-vstream: Efficient real-time understanding for long video streams.
\newblock In \emph{2025 IEEE/CVF International Conference on Computer Vision (ICCV)}, pages 21059--21069. IEEE, 2025{\natexlab{b}}.

\bibitem[Zhang et~al.(2026{\natexlab{a}})Zhang, Yang, Fu, Ng, and Qiu]{zhang2026hermes}
H.~Zhang, S.~Yang, J.~Fu, S.~K. Ng, and X.~Qiu.
\newblock Hermes: Kv cache as hierarchical memory for efficient streaming video understanding.
\newblock In \emph{Proceedings of the 64th Annual Meeting of the Association for Computational Linguistics (Volume 1: Long Papers)}, pages 8411--8430, 2026{\natexlab{a}}.

\bibitem[Zhang et~al.(2026{\natexlab{b}})Zhang, Yang, Wang, Qian, and Xu]{zhang2026querystream}
K.~Zhang, Z.~Yang, B.~Wang, S.~Qian, and C.~Xu.
\newblock Querystream: Advancing streaming video understanding with query-aware pruning and proactive response.
\newblock In \emph{The Fourteenth International Conference on Learning Representations}, 2026{\natexlab{b}}.

\bibitem[Zhang et~al.(2024)Zhang, Wu, Li, Li, Ma, Liu, and Li]{zhang2024llava}
Y.~Zhang, J.~Wu, W.~Li, B.~Li, Z.~Ma, Z.~Liu, and C.~Li.
\newblock Llava-video: Video instruction tuning with synthetic data.
\newblock \emph{arXiv preprint arXiv:2410.02713}, 2024.

\end{thebibliography}
\clearpage
\newpage

\appendix

\section{Additional Experimental Results}

\subsection{Effect of Memory Serialization}
\label{sec:memory_serialization_ablation}

We compare three ways of presenting the same selected visual evidence to the VLM. The \texttt{Default} template preserves memory-block boundaries using neutral numbered segments and token counts. The \texttt{Naive} template instead concatenates all selected visual tokens into a single sequence, whereas the \texttt{Detailed} template exposes both the memory type and its temporal range. All variants use identical selected embeddings, segment order, and readout budget; they differ only in the textual scaffold. The three templates are presented in Figure~\ref{fig:memory_template_variants}.

\begin{figure}[t]
\centering

\begin{tcolorbox}[
    title=\textbf{\texttt{Default}},
    colback=blue!3,
    colframe=black!30,
    boxrule=0.5pt,
    arc=1mm,
    fontupper=\small\ttfamily,
    before upper={\raggedright},
]
    Selected visual memory segments:

    \par\vspace{3pt}

    Memory segment 1 ($n_1$ tokens):
    \par
    $\langle\text{visual token embeddings}_1\rangle$

    \par\vspace{3pt}

    Memory segment 2 ($n_2$ tokens):
    \par
    $\langle\text{visual token embeddings}_2\rangle$

    \par
    \hspace{1em}$\cdots$

    \par\vspace{3pt}

    Question: $\langle\text{Question }q\rangle$
    \end{tcolorbox}

\begin{tcolorbox}[
    title=\textbf{\texttt{Naive}},
    colback=blue!3,
    colframe=black!30,
    boxrule=0.5pt,
    arc=1mm,
    fontupper=\small\ttfamily,
    before upper={\raggedright},
]
<|video|><|video|> $\cdots$ <|video|>

\par\vspace{3pt}

Question: $\langle\text{Question }q\rangle$
\end{tcolorbox}


\begin{tcolorbox}[
    title=\textbf{\texttt{Detailed}},
    colback=blue!3,
    colframe=black!30,
    boxrule=0.5pt,
    arc=1mm,
    fontupper=\small\ttfamily,
    before upper={\raggedright},
]
Selected visual memory segments.

\par\vspace{3pt}

Retrieved key event around
$\langle\min(\boldsymbol{\tau}_1),\max(\boldsymbol{\tau}_1)\rangle$:

\par
<|video|> $\cdots$ <|video|>

\par\vspace{3pt}

Recent context around $\langle T-n,T\rangle$:

\par
<|video|> $\cdots$ <|video|>

\par\vspace{3pt}

Question: $\langle\text{Question }q\rangle$
\end{tcolorbox}
\vspace{-0.1in}
\caption{\textbf{Memory-serialization templates.}
\texttt{Default} preserves memory boundaries using neutral segment labels;
\texttt{Naive} removes these boundaries; and \texttt{Detailed} additionally
exposes memory types and temporal ranges.} \vspace{-0.2in}
\label{fig:memory_template_variants}
\end{figure}

Table~\ref{tab:memory_template_ablation} shows that the default template generally achieves the best performance across the evaluated settings. It supports the importance of preserving memory organization during presentation to the language model. The \texttt{Naive} variant follows the flat presentation commonly adopted by existing token-selection methods, which directly concatenate all retained visual embeddings before language-model decoding. This removes the boundaries between different events and the recent cache, leaving the language model with an unstructured token sequence and making it difficult to distinguish evidence originating from different temporal contexts. The \texttt{Detailed} variant preserves these boundaries but additionally assigns descriptions such as ``key event'' and ``recent context.'' Such labels introduce semantic priors that may bias how the language model weighs the evidence; for example, the word ``key'' may cause a retrieved segment to be overemphasized regardless of its relevance to the current question. In contrast, the \texttt{Default} template preserves each memory block as a separate segment while using only neutral labels and token counts. These results suggest that the benefit comes from preserving the structural organization of memory while avoiding additional semantic cues that may bias how the language model interprets the retained evidence.

\begin{table}[t]
\centering
\small
\setlength{\tabcolsep}{3.5pt}
\renewcommand{\arraystretch}{1.06}
\begin{tabular}{@{}llccc@{}}
\toprule
\textbf{Model} & \textbf{Template} & \textbf{OVO-B} & \textbf{OVO-R} & \textbf{LVB} \\
\midrule
\multirow{3}{*}{Gemma 4 E4B}
& \texttt{naive}    & \textbf{56.40} & 63.72 & 41.21 \\
& \texttt{detailed} & 53.10 & 63.78 & 44.43 \\
& \texttt{default}  & 54.45 & \textbf{64.81} & \textbf{45.77} \\
\midrule
\multirow{3}{*}{NEO-ov 2B}
& \texttt{naive}    & 38.78 & 62.01 & 48.84 \\
& \texttt{detailed} & 39.86 & 61.90 & 48.99 \\
& \texttt{default}  & \textbf{41.90} & \textbf{62.78} & \textbf{49.29} \\
\bottomrule
\end{tabular}
\caption{\textbf{Ablation of memory serialization.} All variants use identical selected visual evidence and differ only in their textual scaffold. OVO-B and OVO-R denote the Backward and Realtime subsets of OVO-Bench, and LVB denotes LongVideoBench. Bold indicates the best result for each model and benchmark.}
\label{tab:memory_template_ablation}
\end{table}

\subsection{Sensitivity to Event Granularity}
\label{sec:event_granularity}

We study how the granularity of each event affects memory quality under a fixed storage budget. Specifically, we vary the per-event visual token size \(m\) and adjust the event-bank capacity \(K\) inversely, keeping \(Km=256\) on OVO-Bench and \(Km=512\) on LongVideoBench. The recent-cache capacity and total readout budget remain unchanged. We also adjust the number of retrieved events so that every variant receives the same recent/event token allocation for each router state. The comparison therefore isolates how the fixed event-memory budget is organized, rather than how many tokens are stored or presented to the language model.

\begin{table}[t]
\centering
\small
\setlength{\tabcolsep}{4pt}
\renewcommand{\arraystretch}{1.08}
\begin{tabular}{@{}ccccc@{}}
\toprule
\(\boldsymbol{m}\)
& \(\boldsymbol{K_{\mathrm{OVO}}}\)
& \textbf{OVO Backward}
& \(\boldsymbol{K_{\mathrm{LVB}}}\)
& \textbf{LongVideoBench} \\
\midrule
8  & 32 & 53.31 & 64 & 44.88 \\
16 & 16 & \textbf{54.45} & 32 & \textbf{45.77} \\
32 & 8  & 50.43 & 16 & 44.88 \\
\bottomrule
\end{tabular}
\caption{\textbf{Event granularity under fixed storage with Gemma 4 E4B.} We keep the total event capacity \(Km\) and the recent/event readout allocation fixed while varying the number and visual token size of stored events.}
\label{tab:event_granularity}
\end{table}

Table~\ref{tab:event_granularity} shows that the intermediate visual token size \(m=16\) performs best on both benchmarks. With \(m=8\), the memory can retain more events, but each event contains less visual evidence and may provide an incomplete representation of its underlying observation. Increasing the visual token budget to \(m=32\) preserves more detail within each event but substantially reduces the number of distinct events that fit within the same storage budget, particularly harming backward reasoning on OVO-Bench. The default setting therefore provides a favorable balance between event-level evidence completeness and temporal coverage. More broadly, the result shows that KeyRec benefits from organizing its bounded storage into moderately sized evidence units rather than maximizing either the number or the individual resolution of stored events.

\subsection{Robustness to Consolidation Thresholds}
\label{sec:merge_sensitivity}

We evaluate the sensitivity of event consolidation to its semantic and temporal thresholds on OVO-Bench Backward. We vary the route-similarity threshold over \(\gamma\in\{0.65,0.75,0.85\}\) and the maximum temporal gap over \(\delta\in\{1,2,4\}\) seconds, while keeping all other settings fixed.

\begin{table}[t]
\centering
\small
\setlength{\tabcolsep}{7pt}
\renewcommand{\arraystretch}{1.08}
\begin{tabular}{@{}lccc@{}}
\toprule
\(\boldsymbol{\gamma\backslash\delta}\)
& \(\boldsymbol{1\,\mathrm{s}}\)
& \(\boldsymbol{2\,\mathrm{s}}\)
& \(\boldsymbol{4\,\mathrm{s}}\) \\
\midrule
0.65 & 54.48 & \textbf{54.67} & 54.27 \\
0.75 & 54.48 & \(54.45^{\dagger}\) & 53.64 \\
0.85 & 54.36 & 53.48 & 52.47 \\
\bottomrule
\end{tabular}
\caption{\textbf{Sensitivity to event-consolidation thresholds on OVO-Bench Backward with Gemma 4 E4B.} \(\gamma\) controls route similarity and \(\delta\) controls temporal locality. \(\dagger\) denotes the default configuration used in the main experiments.}
\label{tab:merge_sensitivity}
\end{table}

As shown in Table~\ref{tab:merge_sensitivity}, KeyRec is stable across a broad neighborhood around the default configuration. The default setting is within \(0.22\) points of the best result, and several nearby threshold combinations obtain comparable accuracy without benchmark-specific tuning. Performance decreases more noticeably when a strict similarity criterion is combined with a wide temporal window. This configuration suppresses the consolidation of moderately changing local observations while still allowing temporally distant near-duplicates to be merged, potentially obscuring the distinction between recurring events. Overall, the results indicate that KeyRec does not depend on a narrowly tuned threshold pair, while also supporting the use of both semantic similarity and temporal locality.

\subsection{Router Behavior Across Backbones}
\label{sec:router_distribution}

\begin{figure}[t]
    \centering
    \includegraphics[width=\linewidth]{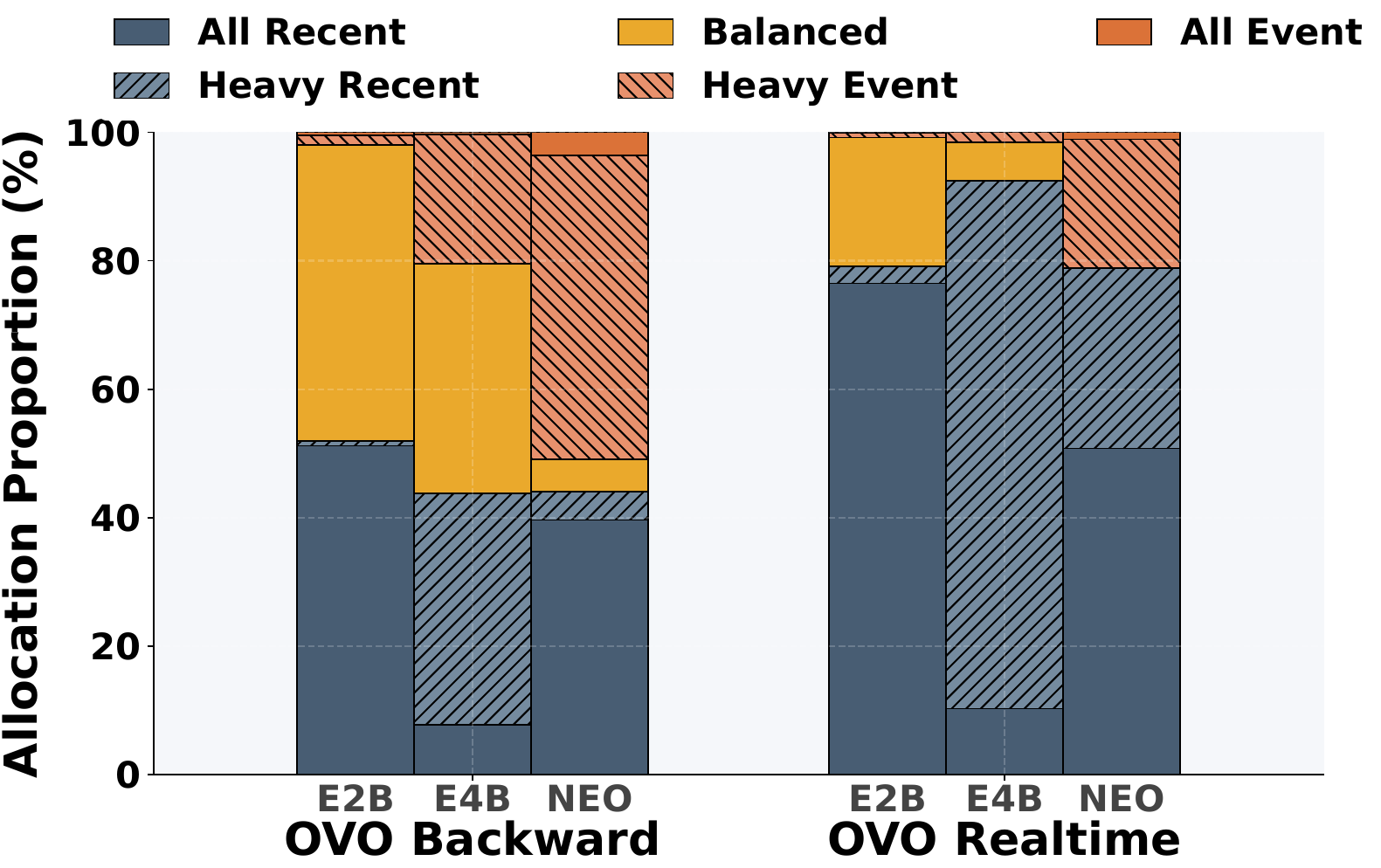} 
    \caption{\textbf{Router-state distributions across backbones on OVO-Bench.} Each stacked bar shows the percentage of questions assigned to each state. Across backbones, realtime questions favor recent memory, while backward questions shift toward balanced or event-oriented ones.}
    \label{fig:router_distribution}
\end{figure}



We examine how the text-only LLM router allocates memory across backbones. Figure~\ref{fig:router_distribution} reports the five routing-state distributions on the OVO-Bench Backward and Realtime subsets. The fine-grained distributions are model-dependent: Gemma 4 E2B favors \textsc{AllRecent} and \textsc{Balanced}, Gemma 4 E4B more frequently selects \textsc{HeavyRecent}, and NEO-2B assigns more backward questions to \textsc{HeavyEvent}.

Despite these differences, a consistent task-dependent shift emerges at the level of memory orientation. For every backbone, realtime questions are routed predominantly toward \textsc{AllRecent} or \textsc{HeavyRecent}. Backward questions shift away from these recency-oriented states toward balanced or event-oriented allocations. Thus, although the models differ in their preferred allocation granularity, they consistently assign more recent evidence to realtime understanding and more historical evidence to backward reasoning. This behavior supports query-adaptive allocation rather than applying a single recent--event split to all questions.

\section{Additional Implementation Details}
\label{sec:additional_implementation}

We use two fixed operating profiles: OVO-Bench and StreamingBench share the \emph{streaming} profile, while LongVideoBench and Video-MME-v2 share the \emph{long-video} profile. No configuration is tuned separately for individual benchmarks.

\paragraph{Memory configuration.}
The streaming and long-video profiles process at most $F=32$ and $F=64$ frames, respectively. Given $n$ model-facing visual tokens per frame, we set the decoder-facing budget to $B=0.1Fn$ and the recent-cache capacity to $N_C=B$. Gemma produces $n=70$ tokens per frame and uses $m=16$ tokens per event, whereas NEO-ov produces approximately $n=210$ tokens per frame and uses $m=48$. The event-bank capacity is $K=16$ for streaming and $K=32$ for long-video evaluation. Thus, for example, the Gemma streaming profile stores at most $224$ recent tokens and $256$ event tokens, totaling $480/2240\approx21\%$ of the dense visual input, while only the $10\%$ readout is exposed to the VLM for each query.

Table~\ref{tab:event_memory_hyperparameters} summarizes the event-memory configuration. In streaming evaluation, temporally adjacent candidates may be consolidated when they satisfy the semantic and temporal criteria. Under the sparse long-video sampling protocol, candidates could be separated beyond the merge window and therefore remain distinct. Temporal coverage is enabled for streaming video to prevent the bounded event bank from concentrating on a narrow portion of an unknown observation horizon. For offline long-video evaluation, frames are sampled across the complete known horizon, and event retention is based on visual utility. This regime-specific temporal weighting follows the same principle as CausalMem~\citep{song2026towards}. Each profile is shared across both backbones and all benchmarks within its corresponding setting.

\begin{table}[t]
\centering
\small
\setlength{\tabcolsep}{3pt}
\begin{tabular}{@{}lcl@{}}
\toprule
\textbf{Parameter} & \textbf{Value} & \textbf{Role} \\
\midrule
\(\gamma\) & \(0.75\) & Route-similarity threshold \\
\(\delta\) & \(2.0\,\mathrm{s}\) & Maximum temporal merge gap \\
\(\beta\)  & \(0.2\) & Route update weight \\
\(\lambda\)& \(0.2\)/\(0.0\) & Temporal-coverage weight \\
\bottomrule
\end{tabular}
\caption{\textbf{Event-memory hyperparameters.} The parameters are shared across settings and backbones. The two values of \(\lambda\) correspond to the streaming and long-video profiles, respectively.}
\label{tab:event_memory_hyperparameters}
\end{table}

\paragraph{Router allocation profiles.}
Table~\ref{tab:router_codebook} reports the effective event-memory allocations induced by Eq.~\ref{eq:budget_allocation}. Because complete events are retrieved, the allocation is discretized into $B/m=14$ event blocks for the streaming profile and $B/m=28$ for the long-video profile. The same codebook is used across backbones within each profile. Figure~\ref{fig:router_prompts} provides the exact text-only router prompts.

\begin{table}[t]
\centering
\small
\setlength{\tabcolsep}{4pt}
\renewcommand{\arraystretch}{1.08}
\begin{tabular}{@{}lcc@{}}
\toprule
\textbf{Router state}
& \textbf{Streaming}
& \textbf{Long video} \\
& \textbf{Fraction (\(k_q\))}
& \textbf{Fraction (\(k_q\))} \\
\midrule
\textsc{AllRecent}   & \(0\;(0)\)     & \(0\;(0)\) \\
\textsc{HeavyRecent} & \(1/7\;(2)\)   & \(2/7\;(8)\) \\
\textsc{Balanced}    & \(2/7\;(4)\)   & \(1/2\;(14)\) \\
\textsc{HeavyEvent}  & \(6/7\;(12)\)  & \(6/7\;(24)\) \\
\textsc{AllEvent}    & \(1\;(14)\)    & \(1\;(28)\) \\
\bottomrule
\end{tabular}
\caption{\textbf{Router allocation codebooks.} Each entry reports the effective fraction of the readout assigned to event memory and the corresponding number of complete events \(k_q\) in parentheses. The same profile is used across backbones.}
\label{tab:router_codebook}\vspace{-0.2in}
\end{table}

\begin{figure*}[t]
\centering

\begin{tcolorbox}[
    title=\textbf{Streaming Video Memory Router Prompt},
    colback=blue!3,
    colframe=black!30,
    boxrule=0.5pt,
    arc=1mm,
    fontupper=\small\ttfamily,
    before upper={\raggedright},
]
You are a memory-allocation router for a streaming video understanding system.

\par\vspace{3pt}

The model can read two kinds of visual memory:

\par
- recent: the latest visual tokens before the question time, useful for current-state and fine visual detail.
\par
- key\_event: selected past visual event tokens, useful for long-range evidence and temporal coverage.

\par\vspace{3pt}

Choose exactly one category for the question:

\par
- all\_recent: the answer should mostly depend on the latest/current visual details.
\par
- heavy\_recent: mostly current-state or fine-detail question, but a little past context may help.
\par
- balanced: both recent context and selected past events are useful.
\par
- heavy\_key\_event: the question likely requires searching across earlier events or broad temporal coverage.
\par
- all\_key\_event: the question is mainly about historical events and recent visual context is unlikely to help.

\par\vspace{3pt}

Return only JSON in this format:

\par
\{"category": "<one category>", "reason": "<short reason>"\}

\par\vspace{3pt}

Question: $\langle\text{Question }q\rangle$
\end{tcolorbox}

\begin{tcolorbox}[
    title=\textbf{Long Video Memory Router Prompt},
    colback=blue!3,
    colframe=black!30,
    boxrule=0.5pt,
    arc=1mm,
    fontupper=\small\ttfamily,
    before upper={\raggedright},
]
You are a memory-allocation router for offline whole-video question answering.

\par\vspace{3pt}

The complete video has already been processed before the question is asked. The model can read two kinds of compressed visual memory:

\par
- recent: visual tokens from only the ending portion of the video. They do not represent the current moment at query time.

\par
- key\_event: selected visual events distributed across the full video timeline.

\par\vspace{3pt}

Choose exactly one category according to where and how broadly the answer evidence is likely to occur:

\par
- all\_recent: use only when the question explicitly asks about the final scene, ending, or final state.

\par
- heavy\_recent: the evidence is likely concentrated near the end, but some earlier context may help.

\par
- balanced: the evidence is localized but its temporal position is uncertain, or both ending and earlier context may matter.

\par
- heavy\_key\_event: the question requires searching, comparing, ordering, counting, or summarizing events across the video.

\par
- all\_key\_event: broad historical coverage is essential and fine detail from the ending is unlikely to help.

\par\vspace{3pt}

Return only JSON in this format:

\par
\{"category": "<one category>", "reason": "<short reason>"\}

\par\vspace{3pt}

Question: $\langle\text{Question }q\rangle$
\end{tcolorbox}

\caption{\textbf{Text-only router prompts.} The streaming router memory prompt interprets recent memory as evidence immediately preceding the query, whereas the offline long video memory router prompt treats it as evidence from the ending portion of the complete video.}
\label{fig:router_prompts}
\end{figure*}

\paragraph{Baseline configurations.}
All baselines use the same frozen backbone, sampled frames, benchmark prompts, and answer-extraction procedure as KeyRec. STC-Pruner uses its Gaussian token-scoring variant. StreamingTOM-CTR uses a temporal-similarity threshold of $0.9$, DPC neighborhood size $7$, and merge weight $0.6$; we evaluate only its causal temporal-reduction component and exclude post-VLM KV-cache quantization. Because it requires attention signals from a dedicated vision encoder, StreamingTOM-CTR is evaluated only on Gemma. CausalMem uses at most $64$ basis vectors, an activity-decay factor of $0.9$, and at most eight new basis vectors per frame; following its original protocol, its temporal weight is $0.8$ for streaming evaluation and disabled for long-video evaluation. All compressed baselines use a $10\%$ decoder-facing retention ratio.

\begin{figure*}[!t]
    \centering
    \includegraphics[width=0.99\textwidth]{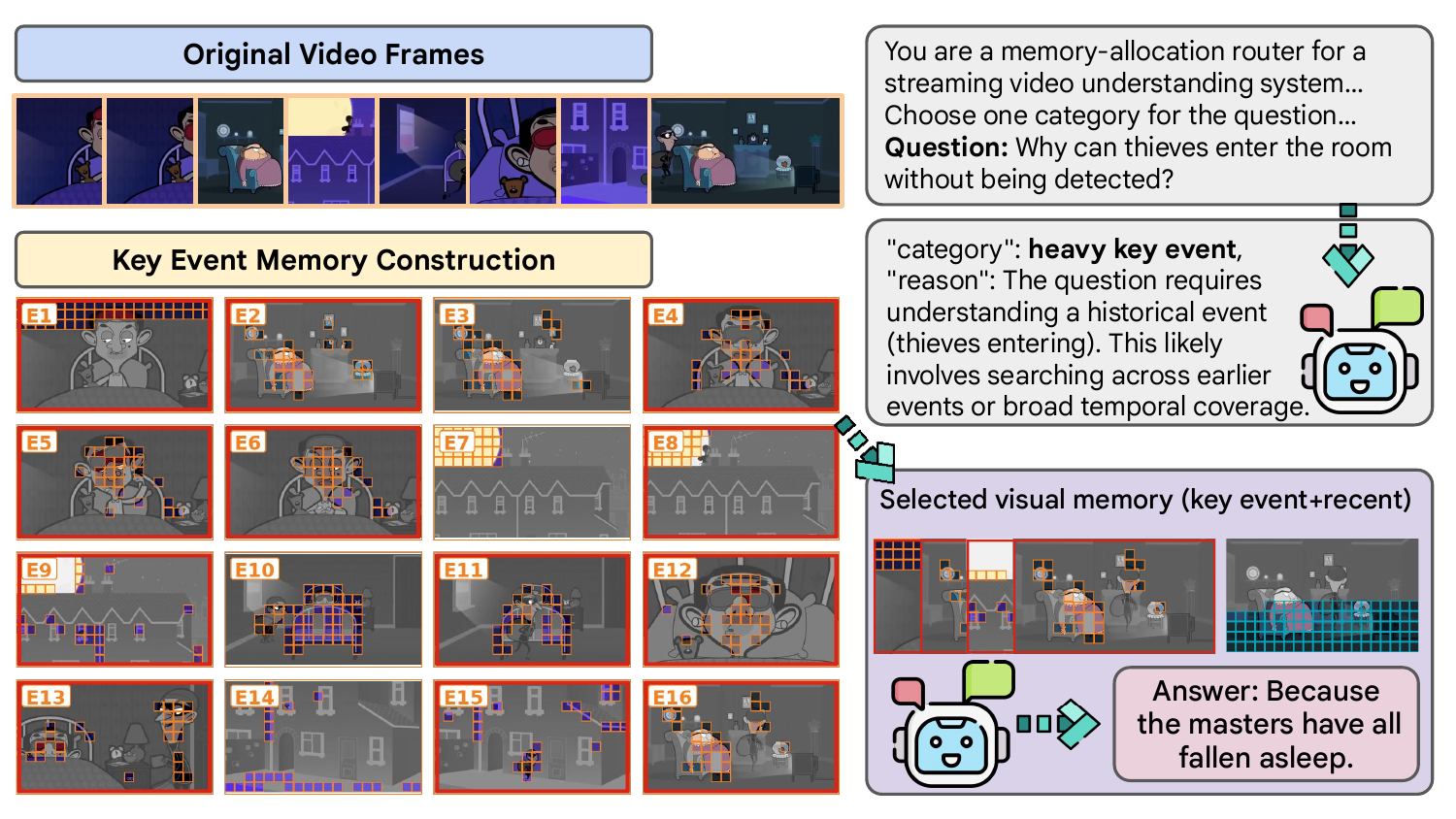} 
    \caption{\textbf{KeyRec case study on StreamingBench with NEO-ov 2B.} KeyRec constructs query-agnostic key events from selected visual patches, routes the history-dependent question to coverage-oriented retrieval, and produces a fixed-budget readout that supports the correct answer. The red border indicates the selected events after determining the budget allocation by the router.}
    \label{fig:case_study}
\end{figure*}

\section{Qualitative Case Study}
\label{sec:case_study}

Figure~\ref{fig:case_study} traces how KeyRec processes a StreamingBench example with NEO-ov 2B. During query-agnostic memory construction, KeyRec retains the visually informative patches highlighted by the orange boxes and organizes them into key events. Because NEO-ov has no separate ViT encoder, these selected patch embeddings are exactly the visual evidence available to its unified backbone at query time, rather than an attribution visualization over additional hidden image features. The resulting events preserve sufficient information about the characters and their states before the question is observed.

The question, \emph{``Why can thieves enter the room without being detected?''}, appears in a streaming benchmark but requires evidence distributed across the preceding history rather than only the latest frame. The LLM router accordingly assigns it to the coverage-oriented heavy-event state. KeyRec then orders the event bank by event-center time and selects events uniformly over the observed history. The bank contains visually similar pairs, such as E2--E3 and E7--E8, which remain separate because their temporal gaps exceed the merge threshold $\delta$. Nevertheless, the fixed-budget readout includes E2 and E8 while excluding E3 and E7, thereby covering both portions of the history without presenting both members of either redundant pair. Together with the recent contribution, this evidence enables NEO-ov to correctly infer that the thieves remain undetected because the masters have fallen asleep.

\section{Per-Task and Category-Level Results}
\label{sec:per_task_results}

Tables~\ref{tab:ovo_per_task_results}--\ref{tab:videomme_v2_official_ratings} provide finer-grained results underlying the aggregate scores reported in Table~\ref{tab:main_results}. On OVO-Bench, KeyRec's gains are most consistent on the real-time tasks, while backward-tracing performance is more mixed across backbones. On StreamingBench, KeyRec achieves the best overall accuracy on all three backbones and performs strongly across most task categories, although some categories such as Cnt remain challenging. On Video-MME-v2, KeyRec is the strongest compressed method across all reported rating dimensions for Gemma 4 E4B and NEO-ov 2B, while the Gemma 4 E2B results are more competitive across methods.

\vspace{-0.1in}
\section{Limitations}
\label{sec:limitations}


KeyRec is training-free and relies on fixed novelty, consolidation, and retention rules; learning these policies from token-level supervision or downstream QA signals may further improve memory construction and routing. In our Gemma implementation, KeyRec operates after the vision encoder, reducing visual-memory and language-model costs but not ViT encoding latency~\cite{kim2026liteframe}. Preliminary intermediate-token reduction did not improve wall-clock encoding time because it disrupted the optimized encoder path, leaving efficient integration of encoder-side pruning with post-encoder memory as an open systems challenge. Finally, the event memory stores selected visual patches without explicitly modeling object identities, trajectories, or state transitions, limiting precise reasoning about duration, order, and cross-frame evolution. Object-centric or hierarchical event representations may better preserve such temporal dynamics under a bounded memory budget.

\FloatBarrier

\begin{table*}[t]
\centering
\scriptsize
\setlength{\tabcolsep}{3pt}
\resizebox{\textwidth}{!}{
\begin{tabular}{llccccccccccc}
\toprule
\multirow{2}{*}{Model} & \multirow{2}{*}{Method} & \multicolumn{4}{c}{Backward} & \multicolumn{7}{c}{Realtime} \\
\cmidrule(lr){3-6}\cmidrule(lr){7-13}
& & EPM & ASI & HLD & Avg. & STU & OJR & ATR & ACR & OCR & FPD & Avg. \\
\midrule
\multirow{5}{*}{Gemma 4 E2B}
& Vanilla
& 41.41 & 39.86 & 48.39 & 43.22
& 36.52 & 44.02 & 58.62 & 44.95
& 63.76 & 56.44 & 50.72 \\ \cmidrule(lr){2-13}
& STC-Pruner
& 36.70 & 35.81 & 51.08 & 41.20
& 41.57 & 42.93 & 56.03 & 35.78
& 56.38 & 55.45 & 48.02 \\
& StreamingTOM-CTR
& 36.36 & \underline{40.54} & 50.00 & 42.30
& 35.96 & 42.39 & 47.41 & \underline{43.12}
& 57.05 & 57.43 & 47.23 \\
& CausalMem
& \textbf{40.40} & \textbf{41.22} & \textbf{54.30} & \textbf{45.31}
& \underline{45.51} & \underline{52.72} & \textbf{62.93}
& 42.20 & \textbf{71.14} & \underline{58.42} & \underline{55.49} \\
& KeyRec
& \underline{37.71} & \underline{40.54} & \underline{53.23} & \underline{43.83}
& \textbf{46.63} & \textbf{54.89} & \underline{62.07}
& \textbf{50.46} & \underline{67.79} & \textbf{64.36} & \textbf{57.70} \\

\midrule
\multirow{5}{*}{Gemma 4 E4B}
& Vanilla
& 45.12 & 59.46 & 56.99 & 53.86
& 52.25 & 52.72 & 66.38 & 54.13
& 71.14 & 62.38 & 59.83 \\ \cmidrule(lr){2-13}
& STC-Pruner
& 37.37 & 50.68 & \textbf{70.97} & \underline{53.01}
& 43.26 & 46.20 & 55.17 & 48.62
& 56.38 & 60.40 & 51.67 \\
& StreamingTOM-CTR
& \textbf{41.41} & 46.62 & 63.44 & 50.49
& 47.19 & 49.46 & 54.31 & 46.79
& 62.42 & 59.41 & 53.26 \\
& CausalMem
& \underline{40.74} & \underline{52.70} & \underline{69.89} & \textbf{54.45}
& \underline{52.81} & \underline{56.52} & \textbf{69.83}
& \underline{58.72} & \underline{73.83} & \underline{63.37}
& \underline{62.51} \\
& KeyRec
& \textbf{41.41} & \textbf{54.73} & 67.20 & \textbf{54.45}
& \textbf{56.74} & \textbf{59.78} & \underline{67.24}
& \textbf{64.22} & \textbf{76.51} & \textbf{64.36}
& \textbf{64.81} \\

\midrule
\multirow{5}{*}{NEO-2B}
& Vanilla
& 48.48 & 54.05 & 13.44 & 38.66
& 44.38 & 54.35 & 69.83
& 51.38 & 67.11 & 61.39 & 58.07 \\ \cmidrule(lr){2-13}
& STC-Pruner
& 39.73 & 35.81 & \textbf{30.11} & 35.22
& 31.46 & 34.24 & 41.38
& 31.19 & 34.23 & 48.51 & 36.84 \\
& StreamingTOM-CTR
& -- & -- & -- & --
& -- & -- & -- & -- & -- & -- & -- \\
& CausalMem
& \underline{41.75} & \underline{45.95} & \underline{29.57}
& \underline{39.09}
& \underline{34.83} & \underline{39.13} & \underline{50.00}
& \underline{36.70} & \underline{50.34} & \underline{55.45}
& \underline{44.41} \\
& KeyRec
& \textbf{48.15} & \textbf{50.68} & 26.88 & \textbf{41.90}
& \textbf{52.25} & \textbf{62.50} & \textbf{73.28}
& \textbf{54.13} & \textbf{71.14} & \textbf{63.37}
& \textbf{62.78} \\
\bottomrule
\end{tabular}
}
\caption{\textbf{Per-task accuracy on the nine OVO-Bench tasks}.
Bold and underline indicate the best and second-best compressed methods, respectively; Vanilla is shown as the uncompressed reference. Backward contains EPM, ASI, and HLD; Realtime contains STU,
OJR, ATR, ACR, OCR, and FPD.}
\label{tab:ovo_per_task_results}
\end{table*}

\begin{table*}[t]
\centering
\scriptsize
\setlength{\tabcolsep}{3pt}
\resizebox{\textwidth}{!}{
\begin{tabular}{llrrrrrrrrrrr}
\toprule
Backbone & Method & CS & OR & AtR & PR & AcR & SU & EU & Cnt & TRU & CR & Overall \\
\midrule
\multirow{5}{*}{Gemma 4 E2B}
& Vanilla & 59.31 & 54.77 & 60.00 & 60.19 & 41.64 & 45.53 & 43.75 & 43.01 & 57.32 & 37.50 & 51.28 \\
\cmidrule(lr){2-13}
& STC-Pruner & 63.41 & 56.87 & 56.39 & 50.93 & 48.73 & 42.04 & 53.75 & \textbf{41.97} & 53.75 & \underline{54.69} & 52.91 \\  
& StreamingTOM-CTR & 62.46 & 55.31 & 56.07 & 40.74 & 47.88 & 46.34 & 51.88 & \underline{32.12} & 53.27 & 53.12 & 51.36 \\
& CausalMem & \underline{67.82} & \underline{62.67} & \underline{68.52} & \underline{53.70} & \underline{53.82} & \underline{56.10} & \underline{57.50} & 31.09 & \underline{58.88} & 53.12 & \underline{58.01} \\
& KeyRec & \textbf{72.24} & \textbf{65.94} & \textbf{74.43} & \textbf{65.74} & \textbf{56.66} & \textbf{57.72} & \textbf{64.38} & 17.62 & \textbf{64.80} & \textbf{58.59} & \textbf{61.29} \\
\midrule
\multirow{5}{*}{Gemma 4 E4B}
& Vanilla & 71.92 & 65.67 & 68.20 & 67.59 & 59.49 & 54.47 & 60.62 & 54.92 & 71.03 & 52.34 & 63.73 \\
\cmidrule(lr){2-13}
& STC-Pruner & 65.30 & 57.69 & 57.70 & \textbf{71.30} & 51.84 & 46.12 & 61.25 & \underline{41.45} & 58.75 & 58.59 & 56.44 \\
& StreamingTOM-CTR & 70.03 & 58.31 & 64.59 & 60.19 & 54.67 & 55.69 & \underline{65.62} & \textbf{51.30} & 60.75 & \underline{60.94} & 60.25 \\
& CausalMem & \underline{71.61} & \underline{66.49} & \underline{72.46} & \underline{70.37} & \underline{58.92} & \underline{61.38} & \underline{65.62} & 37.31 & \underline{65.11} & 59.38 & \underline{63.61} \\
& KeyRec & \textbf{76.66} & \textbf{68.94} & \textbf{78.36} & 68.52 & \textbf{60.91} & \textbf{66.26} & \textbf{66.25} & 40.41 & \textbf{65.73} & \textbf{61.72} & \textbf{66.49} \\
\midrule
\multirow{5}{*}{NEO-ov 2B}
& Vanilla & 70.03 & 76.29 & 77.70 & 74.07 & 66.01 & 59.35 & 68.12 & 52.85 & 67.60 & 77.34 & 69.06 \\
\cmidrule(lr){2-13}
& STC-Pruner & 52.05 & 49.32 & 53.44 & \underline{59.26} & 52.12 & 46.75 & 52.50 & \textbf{32.64} & 52.65 & \underline{60.94} & 50.68 \\
& StreamingTOM-CTR & -- & -- & -- & -- & -- & -- & -- & -- & -- & -- & -- \\
& CausalMem & \underline{59.62} & \underline{56.68} & \underline{60.33} & 57.41 & \underline{54.67} & \underline{52.85} & \underline{65.00} & 29.53 & \underline{57.32} & 60.16 & \underline{55.56} \\
& KeyRec & \textbf{74.76} & \textbf{75.20} & \textbf{81.97} & \textbf{64.81} & \textbf{68.56} & \textbf{68.29} & \textbf{70.00} & \underline{31.61} & \textbf{68.54} & \textbf{70.31} & \textbf{69.10} \\
\bottomrule
\end{tabular}
}
\caption{\textbf{Per-task accuracy on the ten StreamingBench Real-Time tasks}. Bold and underline indicate the best and second-best compressed methods, respectively; Vanilla is shown as the uncompressed reference.}
\label{tab:streamingbench_per_task_accuracy}
\end{table*}

\begin{table*}[t]
\centering
\small
\resizebox{\textwidth}{!}{
\begin{tabular}{llrrrrrrrr}
\toprule
Backbone & Method & Unnorm. Acc. & Level 1 & Level 2 & Level 3 & Relevance & Rel.-Linear & Logic &  Total \\
\midrule
\multirow{5}{*}{Gemma 4 E2B}
& Vanilla
& 18.47 & 10.19 & 6.01 & 4.84
& 7.36 & 17.29 & 5.09 & 6.56 \\ \cmidrule(lr){2-10}
& STC-Pruner
& 14.16 & 4.69 & 4.03 & 3.54
& 4.35 & 12.96 & 3.32 & 3.99 \\
& StreamingTOM-CTR
& \textbf{15.69} & \textbf{6.73} & \textbf{5.77}
& 3.93 & \textbf{5.92} & \textbf{14.93}
& \underline{3.89} & \textbf{5.21} \\
& CausalMem
& 15.31 & 5.37 & 4.51 & \textbf{4.49}
& \underline{5.24} & \underline{13.82}
& 3.76 & 4.72 \\
& KeyRec
& \underline{15.41} & \underline{5.69} & \underline{4.82}
& \underline{4.35} & \underline{5.24} & 13.73
& \textbf{4.10} & \underline{4.84} \\

\midrule
\multirow{5}{*}{Gemma 4 E4B}
& Vanilla
& 20.09 & 10.44 & 7.56 & 6.03
& 8.25 & 18.55 & 6.47 & 7.62 \\ \cmidrule(lr){2-10}
& STC-Pruner
& 14.34 & 5.29 & 5.47 & 3.79
& 5.31 & 13.15 & 3.55 & 4.69 \\
& StreamingTOM-CTR
& \underline{15.91} & \underline{6.75} & \underline{5.93}
& \underline{4.56} & \underline{6.05} & \underline{14.55}
& \underline{4.61} & \underline{5.54} \\
& CausalMem
& 14.84 & 5.02 & 5.53 & 3.91
& 5.13 & 13.01 & 3.89 & 4.70 \\
& KeyRec
& \textbf{17.63} & \textbf{6.97} & \textbf{6.05}
& \textbf{5.19} & \textbf{6.19} & \textbf{15.41}
& \textbf{5.39} & \textbf{5.91} \\

\midrule
\multirow{5}{*}{NEO-ov 2B}
& Vanilla
& 22.06 & 13.18 & 8.18 & 6.55
& 9.85 & 21.00 & 6.68 & 8.74 \\ \cmidrule(lr){2-10}
& STC-Pruner
& \underline{17.56} & \underline{7.57} & \underline{5.98}
& 4.59 & \underline{7.50} & \underline{17.97}
& 2.59 & \underline{5.78} \\
& StreamingTOM-CTR
& -- & -- & -- & --
& -- & -- & -- & -- \\
& CausalMem
& 17.47 & 6.52 & 5.79 & \underline{4.63}
& 6.54 & 16.71 & \underline{3.49} & 5.47 \\
& KeyRec
& \textbf{20.03} & \textbf{10.30} & \textbf{6.68}
& \textbf{4.83} & \textbf{7.86} & \textbf{19.22}
& \textbf{4.81} & \textbf{6.79} \\
\bottomrule
\end{tabular}
}
\caption{\textbf{Video-MME v2 official ratings and unnormalized accuracy}. Bold and underline indicate the best and second-best compressed methods, respectively; Vanilla is shown as the uncompressed reference.}
\label{tab:videomme_v2_official_ratings}
\end{table*}

\end{document}